%% file: main.tex
\documentclass[10pt,twocolumn,letterpaper]{article}

\usepackage[pagenumbers]{cvpr} 

\input{preamble}
\definecolor{cvprblue}{rgb}{0.21,0.49,0.74}
\usepackage[pagebackref,breaklinks,colorlinks,allcolors=cvprblue]{hyperref}
\usepackage{multirow}
\def\paperID{} 
\def\confName{CVPR}
\def\confYear{2026}

\title{CoT-Seg: Rethinking Segmentation with\\
Chain-of-Thought Reasoning and Self-Correction}

\author{\vspace{0.5em}
 Shiu-hong Kao$^{1,\dagger}$ \hspace{10pt} Chak Ho Huang$^{1,\dagger}$ 
 \hspace{10pt} Huaiqian Liu$^{1,\dagger}$ 
 \hspace{10pt} Yu-Wing Tai$^2$ \hspace{10pt} Chi-Keung Tang$^1$ \\
 {\small $^1$The Hong Kong University of Science and Technology, $^2$Dartmouth College}\\
 \vspace{0.5em}
 {\small $^\dagger$\textit{Equal contribution}}
}

\begin{document}
\maketitle
 \input{sec/0_abstract}    
 \input{sec/1_intro}
 \input{sec/2_related_work}

\input{sec/3_method}

\input{sec/4_experiment}

\input{sec/5_conclusion}
 {
    \small
    \bibliographystyle{ieeenat_fullname}
    \bibliography{main}
 }
\input{sec/6_Appendix}

\end{document}

%% file: sec/0_abstract.tex
\begin{abstract}
Existing works of reasoning segmentation often fall short in complex cases, particularly when addressing complicated queries and out-of-domain images. Inspired by the chain-of-thought reasoning, where harder problems require longer thinking steps/time, this paper aims to explore a system that can think step-by-step, look up information if needed, generate results, self-evaluate its own results, and refine the results, in the same way humans approach harder questions. We introduce \textbf{CoT-Seg}, a training-free framework that rethinks reasoning segmentation by combining \textbf{chain-of-thought reasoning} with \textbf{self-correction}. Instead of fine-tuning, CoT-Seg leverages the inherent reasoning ability of pre-trained MLLMs (\textit{e.g.,} GPT-4o) to decompose queries into meta-instructions, extract fine-grained semantics from images, and identify target objects even under implicit or complex prompts. Moreover, CoT-Seg incorporates a self-correction stage: the model evaluates its own segmentation against the original query and reasoning trace, identifies mismatches, and iteratively refines the mask. This tight integration of reasoning and correction significantly improves reliability and robustness, especially in ambiguous or error-prone cases. Furthermore, our CoT-Seg  framework allows easy incorporation of  retrieval-augmented reasoning, enabling the system to access external knowledge when the input lacks sufficient information. To showcase CoT-Seg's ability to handle very challenging cases, we introduce a new dataset {\sc ReasonSeg-Hard}. Our results highlight that combining chain-of-thought reasoning, self-correction,  offers a powerful paradigm for vision language integration driven segmentation. Our project website is available at \url{https://danielshkao.github.io/cot-seg.html}.
\end{abstract}

%% file: sec/1_intro.tex
\section{Introduction}
\label{sec:intro}
\begin{figure*}[h]
    \centering
    \includegraphics[width=0.9\linewidth]{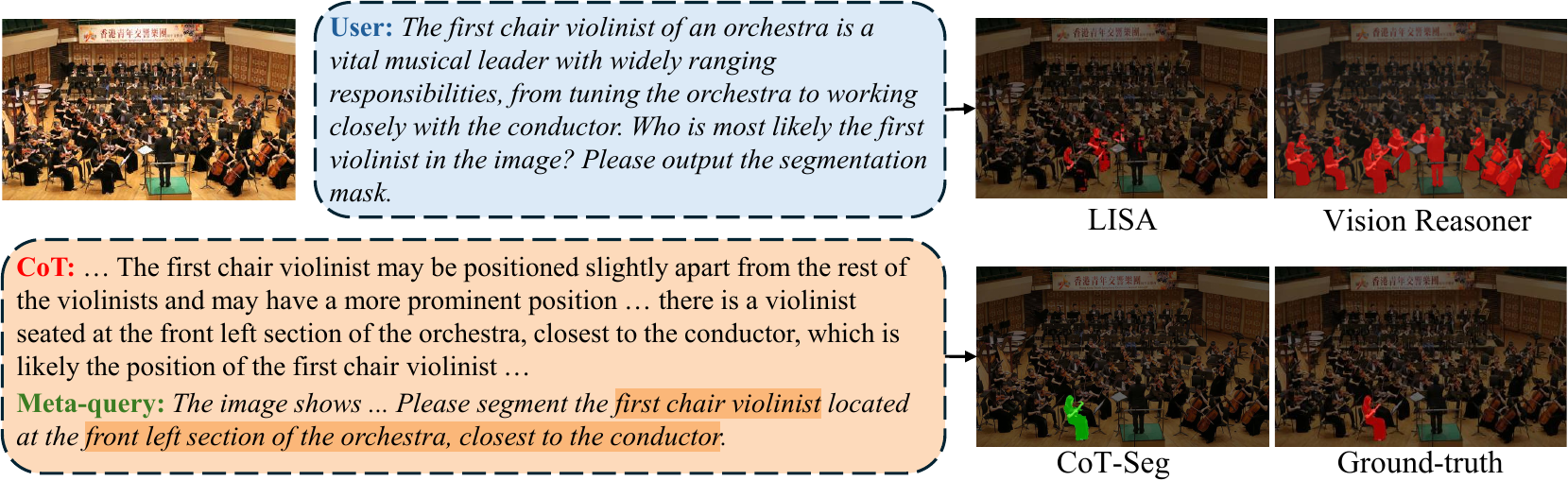}
    \vspace{-0.1in}
    \caption{Finding the first violinist (concertmaster) is challenging among similar-looking musicians. CoT-Seg reasons that they sit to the conductor's left and generates a meta-query with relevant {\em spatial} information, enabling more accurate segmentation than LISA and Vision Reasoner (No self-correction was needed). }
    \vspace{-0.15in}\label{fig:violinist}
\end{figure*}
\begin{figure*}[h]
    \centering
    \includegraphics[width=0.9\linewidth]{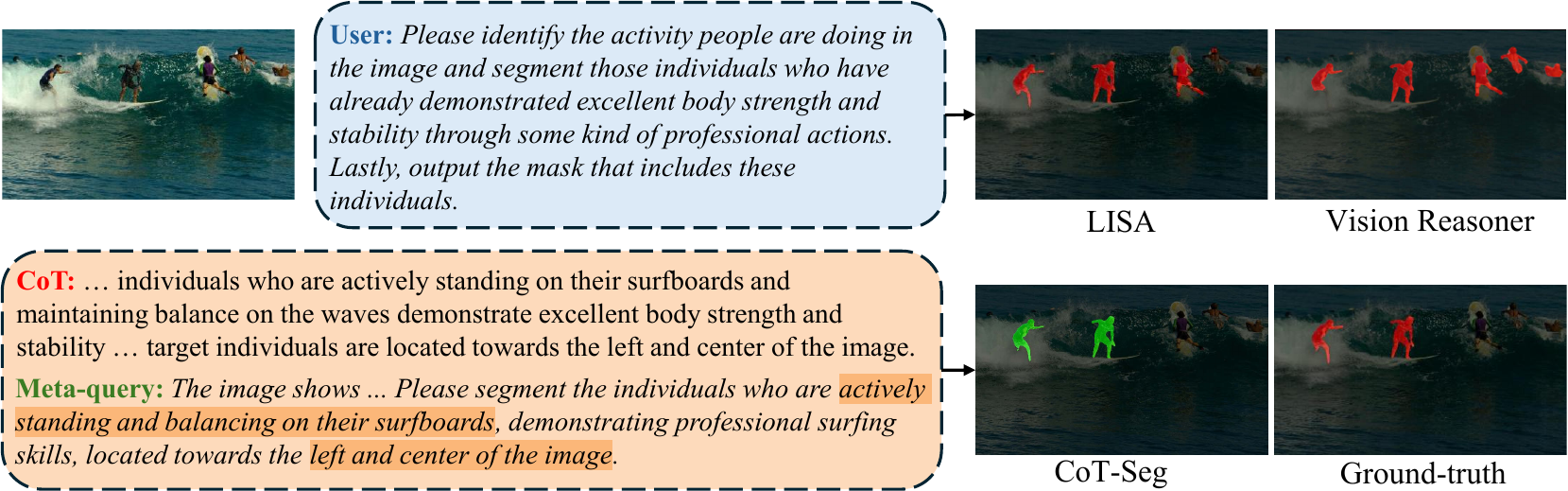}
    \vspace{-0.1in}
    \caption{CoT-Seg reasons about the user’s query to segment surfers in the correct {\em pose}, capturing only those who have popped up and are riding waves, unlike LISA and Vision Reasoner (No self-correction was needed).}
    \vspace{-0.15in}
    \label{fig:surfers}
\end{figure*}
\begin{figure*}[h]
    \centering
    \includegraphics[width=0.9\linewidth]{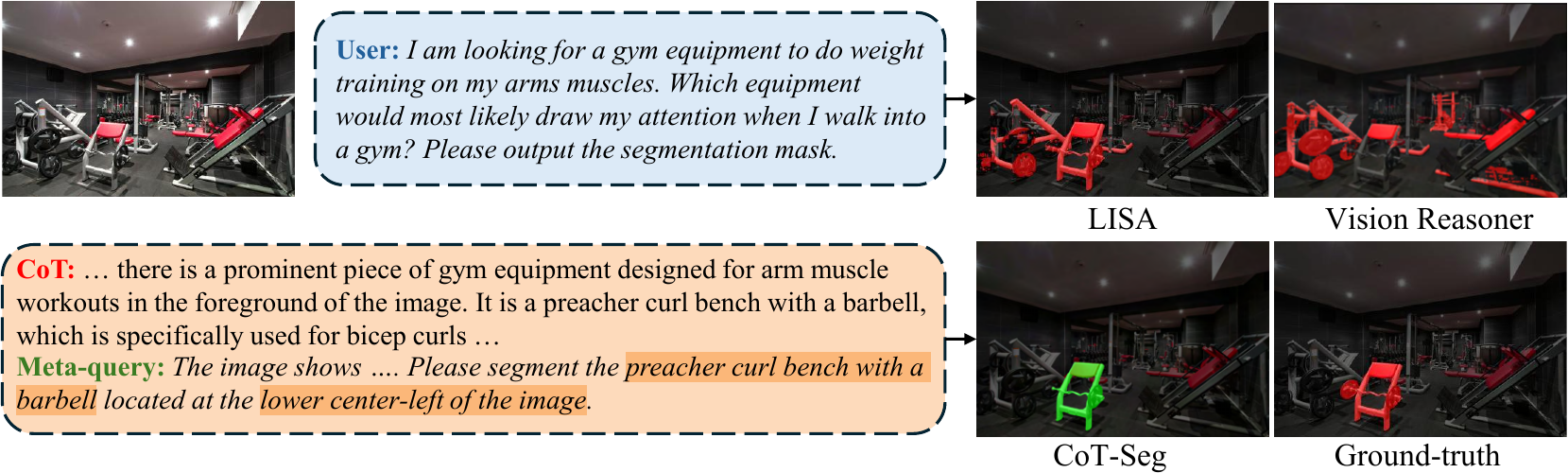}
    \vspace{-0.15in}
    \caption{CoT-Seg identifies the gym equipment matching the user's query for biceps, e.g., the preacher’s curl, reasoning about its {\em function} without any training (Self-correction was needed).}
    \vspace{-0.15in}
    \label{fig:bicep}
\end{figure*}
\begin{figure*}[h]
    \centering
    \includegraphics[width=0.9\linewidth]{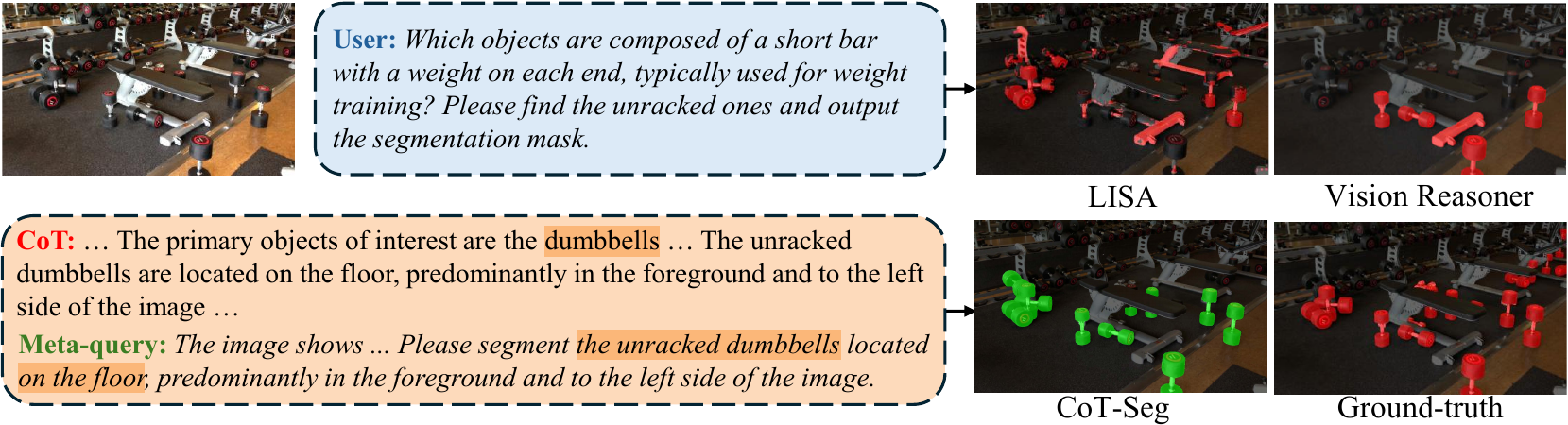}
    \vspace{-0.15in}
    \caption{CoT-Seg reasons about the arrangement of dumbbells to segment those that are {\em unracked}, a more challenging task than simple detection (No self-correction was needed). }
    \label{fig:unracked}
    \vspace{-0.15in}
\end{figure*}

Reasoning segmentation represents a promising step toward vision-language integration, where a system generates a segmentation mask from complex and often implicit language queries. Recent progress has been driven by fine-tuning Multimodal Large Language Models (MLLMs), such as LISA~\citep{lai2023lisa}, Seg-Zero~\citep{liu2025seg} and Vision Reasoner ~\citep{liu2025visionreasoner}, to produce segmentation outputs. Despite their success, these methods struggle with cases that require nuanced reasoning, domain knowledge, or contextual inference which are the major challenges that humans naturally handle.

Consider the examples in Figures~\ref{fig:violinist}--\ref{fig:unracked}. Locating the first-chair violinist requires knowledge of orchestra seating arrangements, not just visual similarity. Differentiating surfers by posture demands reasoning about dynamic body positions. Selecting the correct gym equipment for bicep training requires understanding functional affordances. Identifying unracked dumbbells requires contextual analysis of their relation to the rack.

\textit{How can we approach robust reasoning segmentation?} To improve the segmentation quality, existing work mainly focus on the techniques to connect MLLM with vision foundation model, \textit{e.g.} LISA introduces a new [SEG] token in MLLM vocabulary to extract segmentation-based features, Seg-Zero utilizes reinforcement learning to finetune MLLM for point and bounding box outputs. However, generating difficult mask from implicit text in one pass is even challenging for human beings. We rethink the task, aiming to explore a system capable of approaching the prediction step-by-step. This system allows the model to address harder problems with longer reasoning steps/time. To achieve this, we aim to propose a coherent mechanism to analyze implicit semantics in depth, evaluate its own predictions, and refine its own mistakes.

In this work, we introduce \textbf{CoT-Seg}, a training-free framework that revisits reasoning segmentation by integrating \textit{chain-of-thought (CoT) reasoning} with a dedicated \textit{self-correction mechanism}. Although CoT has been used in other reasoning tasks, to the best of our knowledge it has not been explored for reasoning-driven segmentation, where the system must jointly interpret linguistic instructions, visual context, and object relationships. Designing a CoT workflow in this setting is non-trivial, since the reasoning must both decompose the query and reveal the semantic structure required for spatially grounded segmentation. CoT-Seg leverages the latent reasoning capability of pre-trained MLLMs (e.g., GPT-4o) to convert queries into meta-instructions, extract fine-grained semantics, and generate initial segmentation maps, all without additional training. Crucially, CoT-Seg introduces a self-correction stage: the model checks its predictions against the query and reasoning trace, detects inconsistencies, and refines the results through automatically generated meta-queries. This closed-loop design enables the system not only to reason about segmentation but also to critique and repair its outputs, showing that CoT becomes genuinely impactful when adapted to the challenges of reasoning-based segmentation.


Furthermore, we extend CoT-Seg with {\em retrieval-augmented reasoning}. When the query and image lack sufficient information, CoT-Seg calls an external agent to retrieve relevant knowledge from the web, integrating it into the reasoning process. This augmentation further strengthens its ability to tackle ambiguous or knowledge-intensive cases.

Through extensive experiments on ReasonSeg~\cite{lai2023lisa}, we demonstrate that CoT-Seg outperforms existing methods while requiring no additional training. To examine our work in extremely difficult cases, we also propose a new reasoning segmentation benchmark called {\sc ReasonSeg-Hard}, where CoT-Seg showcases additional improvements against state of the arts. Our results show that integrating CoT reasoning, self-correction, and retrieval augmentation provides a powerful paradigm for advancing reasoning-driven segmentation toward human-level reliability.

%% file: sec/2_related_work.tex
\section{Related Work}\label{sec:related_work}

\noindent{\bf Image Segmentation and Reasoning Segmentation.}  
Image segmentation has evolved from early graphical-model-based methods, such as Conditional Random Fields (CRFs)~\citep{krahenbuhl2011efficient,chen2017deeplab} and region growing~\citep{dias2019semantic}, to deep learning approaches that utilize encoder-decoder architectures~\citep{badrinarayanan2017segnet}, dilated convolutions~\citep{yu2015multi}, pyramid pooling~\citep{zhao2017pyramid}, and non-local operators~\citep{liu2015parsenet}. Instance segmentation~\citep{he2017mask,cheng2022masked} and panoptic segmentation~\citep{kirillov2019panoptic,cheng2020panoptic} further pushed the boundary to finer-grained understanding.  

The emergence of foundation models for segmentation, especially the Segment Anything Model (SAM)~\citep{kirillov2023segment}, has revolutionized the field. By training on billions of masks and images, SAM enables promptable, zero-shot segmentation with multimodal inputs like points or bounding boxes. Leveraging SAM with Multimodal Large Language Models (MLLMs) has led to a new line of works on reasoning segmentation~\citep{lai2024lisa,xia2024gsva,zhang2023next,he2024multi,yao2025mmreasonopenendedmultimodalmultistep}. These approaches generate segmentation masks conditioned on implicit or complex textual queries. However, combining MLLMs with SAM directly often fails in challenging scenarios, such as queries requiring domain knowledge, occluded objects, or intricate structures. In contrast, our work shows that \textit{integrating chain-of-thought reasoning and self-correction} can substantially enhance robustness and accuracy in these difficult cases.  

\noindent{\bf Chain-of-Thought Reasoning in LLMs and MLLMs.}  
Chain-of-Thought (CoT) reasoning improves reasoning performance in large language models by decomposing complex tasks into intermediate steps~\citep{wei2022chain,wang2022self,zhang2022automaticchainthoughtprompting,lyu2023faithful,kojima2023largelanguagemodelszeroshot}. While CoT has been extensively explored in text-only LLMs, its integration into Multimodal LLMs (MLLMs) is more challenging. Existing approaches often rely on fine-tuning MLLMs with multimodal CoT datasets~\citep{mondal2024kamcotknowledgeaugmentedmultimodal,zhang2024multimodalchainofthoughtreasoninglanguage,lu2022learn} or introducing intermediate representations like graphs~\citep{mitra2024compositional} or code~\citep{suris2023vipergpt}, which limit accessibility and scalability.  

Recent works highlight the potential of \textit{test-time CoT reasoning} in pre-trained LLMs~\citep{snell2024scaling} and its applications in visual reasoning~\citep{guo2022images,lian2023llm}, robotics~\citep{hu2023look}, and multimodal planning~\citep{yao2025mmreasonopenendedmultimodalmultistep}. Inspired by these trends, our framework leverages carefully designed CoT prompts in a \textit{training-free manner}, enabling MLLMs to reason over images and textual queries, evaluate initial segmentation outputs, and self-correct without additional training.  

\noindent{\bf Self-Correction and Retrieval-Augmented Reasoning.}  
While CoT provides step-by-step reasoning, errors in initial predictions can propagate if unchecked. Recent studies in reasoning with feedback~\citep{zhao2025boostingllmreasoningspontaneous,he2025selfcorrectionrefinementlearningframework} demonstrate that self-evaluation and iterative refinement improve accuracy. Our method explicitly incorporates a \textit{self-correction loop} for reasoning segmentation, allowing the model to detect inconsistencies and refine segmentation masks.  

Furthermore, retrieval-augmented reasoning~\citep{lewis2021retrievalaugmentedgenerationknowledgeintensivenlp,komeili2021internetaugmenteddialoguegeneration} has shown that external knowledge can enhance reasoning when input information is incomplete. CoT-Seg integrates retrieval mechanisms to access relevant knowledge at test time, enabling more robust segmentation under ambiguous or knowledge-intensive queries.  

Overall, our work is positioned at the intersection of \textit{reasoning segmentation, CoT-enabled MLLMs, self-correction, and retrieval augmentation}, combining these advances into a unified, training-free framework that achieves state-of-the-art performance in complex vision-language tasks.

%% file: sec/3_method.tex
\begin{figure*}[h]
    \centering
    \includegraphics[width=0.75\linewidth]{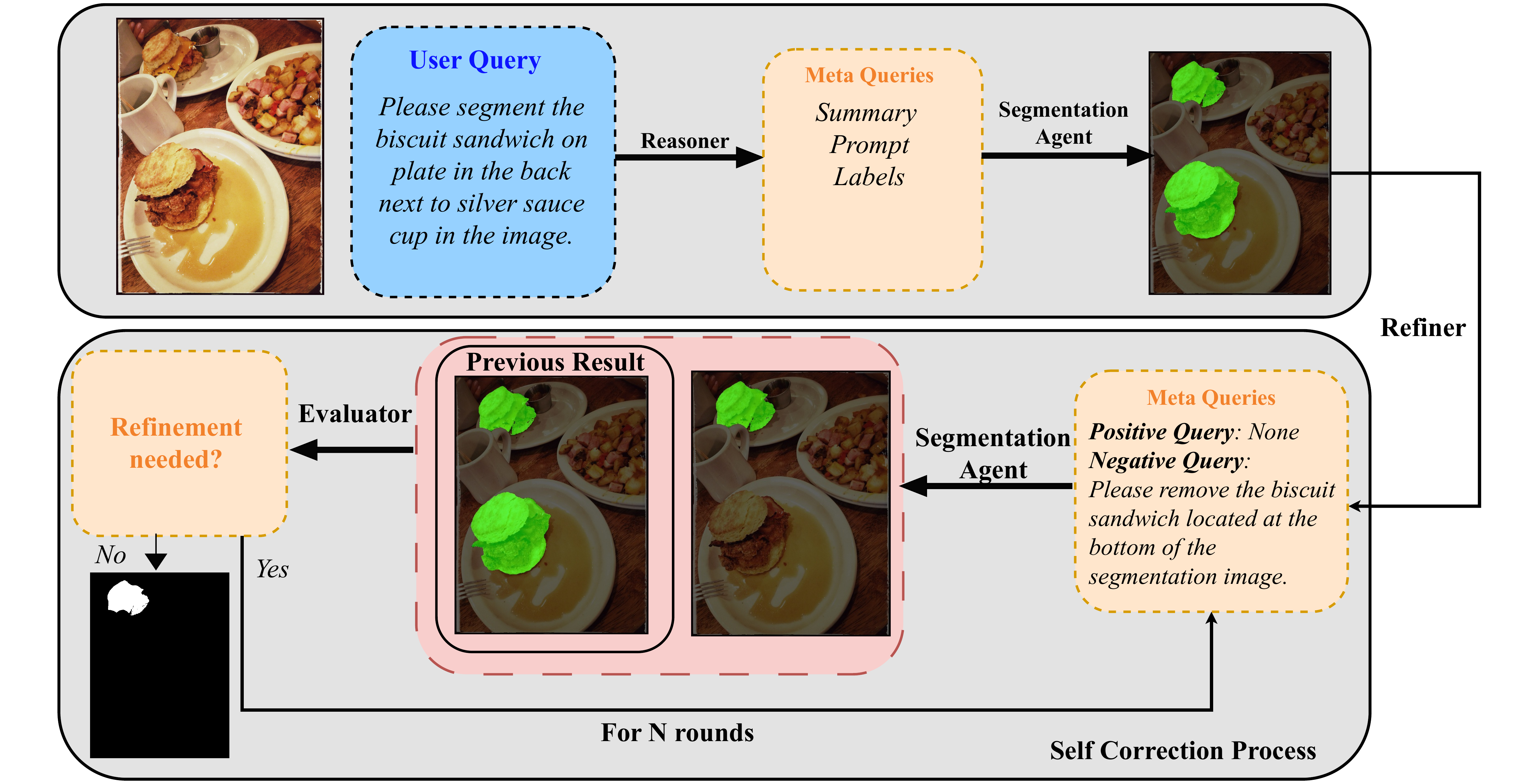}\\
    \vspace{-0.1in}\caption{\textbf{Overview of CoT-Seg.} The pre-trained MLLM Reasoner generates a chain-of-thought (CoT) over the input image and query, producing an explicit meta-query that translates complex, implicit instructions into clear segmentation guidance. The Segmentation Agent predicts the initial mask, which is then optionally refined by the iterative refinement pipeline. The first-turn mask and original image are examined by the MLLM Evaluator which evaluates the mask and decide if any refinement is necessary. If it does require refinement, then it is passed onto the MLLM Refiner which produces two queries to correct for false positives and negatives. These queries are used inline with the segmentation agent to produce a refined mask for the next iteration of refinement.}
    \vspace{-0.15in}
    \label{fig:method}
\end{figure*}

\section{Method}\label{sec:method}

Given an image $I \in \mathbb{R}^{3 \times H \times W}$ and a textual query $q$, reasoning segmentation aims to predict a binary mask $\hat{\mathrm{M}}$ corresponding to the object(s) referred by $q$. CoT-Seg achieves this by combining chain-of-thought reasoning, self-correction, and optional retrieval-augmented reasoning in a multi-agentic framework. The system consists of three collaborating agents: the MLLM \emph{Reasoner}, the \emph{Segmentation Agent}, and the \emph{Evaluator}.

\Cref{fig:method} gives an overview of CoT-Seg, where the Reasoner analyzes the image and query using a chain-of-thought (CoT) process, generating an explicit meta-query that guides the Segmentation Agent. The Segmentation Agent produces an initial mask using the meta-query and its supported input types, such as text, points, bounding boxes, or scribbles. The Evaluator then analyzes the predicted mask in combination with the original query and image, identifying errors and synthesizing refinement meta-queries for self-correction. 
\vspace{-0.05in}
\subsection{MLLM Reasoner}
The Reasoner $\mathcal{R}$ performs step-by-step chain-of-thought (CoT) reasoning to identify the target object(s) in the image. To achieve this, $\mathcal{R}$ utilizes a series of \emph{Question Proposers} that generate questions progressively from coarse to fine. Initially, coarse questions capture high-level scene context and object categories. Based on the answers, subsequent proposers generate finer-grained questions to localize the target objects, reasoning over attributes such as position, size, and relationships with other objects. This iterative process continues until sufficient information is collected to precisely identify the target or until it reaches max number of rounds.  

Formally, each question-answer pair is generated autoregressively:
\begin{equation}
\begin{split}
(Q_k, A_k) = \mathcal{R}(I, q, Q_{<k}, A_{<k}, \text{SegmentorCapabilities}) \\
\quad k = 1, \dots, n    
\end{split}
\end{equation}
where $\text{SegmentorCapabilities}$ is defined as a textual description that informs the Reasoner of which input types the Segmentation Agent supports (e.g., text, points, bounding boxes, scribbles).  

After completing all CoT steps, the Reasoner summarizes the collected information into a structured \emph{meta-query} $\tilde{q}_m$, which is compatible with the Segmentation Agent aligning with SegmentorCapabilities. For non-textual inputs, such as points or scribbles, the meta-query is encoded in a JSON format specifying the input type, coordinates, and spatial attributes:
\begin{equation}
\tilde{q}_m = \mathcal{R}_{\text{summarize}}(\{Q_k, A_k\}_{k=1}^n, \text{SegmentorCapabilities}).
\end{equation}

This structured meta-query is then passed to the Segmentation Agent to produce the initial mask, and subsequently to the Evaluator for self-correction if necessary. By combining coarse-to-fine question proposing with explicit summarization, the Reasoner ensures precise target localization and effective guidance for zero-shot segmentation.

\subsection{Reasoning Segmentation Agent}
The Segmentation Agent $\mathcal{A}$ predicts masks based on the meta-query $\tilde{q}_m$ and its supported input types. It consists of a frozen vision encoder $E$, a mask decoder $\mathcal{D}$, and a vision-language model $\mathcal{F}$ for multimodal encoding~e.g.,~\citep{lai2023lisa,zou2023segment}. The predicted mask is:
\begin{equation}
\hat{\mathrm{M}} = \mathcal{A}(I, \tilde{q}_m) = \mathcal{D}(\mathcal{F}(I, \tilde{q}_m), E(I)).
\end{equation}

By explicitly describing the segmentor’s input capabilities, both the Reasoner and Evaluator can adapt their CoT reasoning. If the segmentation agent cannot support a requested input type, the method may fail, highlighting the dependency on the segmentor’s flexibility. This design ensures that the meta-query generated by the Reasoner is always compatible with the segmentor.

\subsection{Evaluator and Self-Correction}
The Evaluator $\mathcal{J}$ assesses the quality of the mask generated by the Segmentation Agent and guides iterative refinement. It receives the original image $I$, the user query $q$, the predicted mask $\hat{\mathrm{M}}$, and the SegmentorCapabilities as inputs. The Evaluator performs a chain-of-thought (CoT) reasoning process, similar to the Reasoner, to check whether the mask correctly covers the target objects and respects spatial and semantic constraints.  
    
If refinement is needed, the Evaluator generates two types of meta-queries in a structured JSON format: $\tilde{q}_P$ for false negatives and $\tilde{q}_N$ for false positives. These queries specify the type of correction, spatial coordinates, and other relevant control signals compatible with the Segmentation Agent. The segment agent then ouputs binary postive/negative masks $s_P$/$s_N$ which is then added/subtracted to the orignal segmentation. Formally, the refinement process is:
\begin{align}
S &= \mathcal{J}_{\text{assess}}(I, \hat{\mathrm{M}}, q, \text{SegmentorCapabilities}), \\
(\tilde{q}_P, \tilde{q}_N) &= \mathcal{J}_{\text{refine}}(I, \hat{\mathrm{M}}, q, S, \text{SegmentorCapabilities}), \\
s_P &= \mathcal{A}(I, \tilde{q}_P), \quad s_N = \mathcal{A}(I, \tilde{q}_N), \\
s' &= s + s_P - s_N, \\
\hat{\mathrm{M}}' &= \{(i,j) \mid s'_{i,j} > 0\}
\end{align}
where  $s$ implies the prediction score output by the segmentor, satisfying $\hat{\mathrm{M}}=\{s_{i,j}\mid s_{i,j} > \text{threshold}\}$.
This iterative self-correction loop continues until $S=0$ (Correct Segmentation) or a maximum number of refinement rounds is reached. By using structured JSON communication, the Evaluator ensures compatibility with diverse Segmentation Agents and input modalities, enabling robust zero-shot segmentation with automated error correction.  
\textcolor{black}{To ensure that $\hat{\mathrm{M}'}$ does not get worse than
$\hat{\mathrm{M}}$, which may also happen to humans after several refinement turns,  $\mathcal{J}$ will make a judgment whether to revert back to the previous segmentation $\hat{\mathrm{M}}$ as the chosen segmentation.}

\subsection{Multimodal Input Control}
CoT-Seg supports diverse image-based controls in addition to textual queries, including points, bounding boxes, scribbles, and highlighted regions. The Reasoner $\mathcal{R}$ is aware of the Segmentation Agent’s capabilities through the SegmentorCapabilities input. For non-textual inputs, it encodes the meta-query in JSON format specifying input type, coordinates, and spatial attributes. This allows both the Reasoner and Evaluator to generate compatible guidance and refinement instructions.

Given an image $I$ and a control image $I_{ann}$, the Reasoner generates step-by-step CoT reasoning to interpret annotated regions and produce a meta-query $\tilde{q}_m$:
\begin{equation}
\begin{split}
\tilde{q}_m = \mathcal{R}_{\text{summarize}}(\{Q_k, A_k\}_{k=1}^n, \\\text{SegmentorCapabilities}, I_{ann})
\end{split}
\end{equation}
which is then passed to the Segmentation Agent to produce the mask $\hat{\mathrm{M}} = \mathcal{A}(I, \tilde{q}_m)$. The Evaluator can further refine the output via self-correction if necessary, using the same JSON format for multimodal control information.  

\subsection{Retrieval-Augmented Reasoning }
In cases where the input image and query do not provide sufficient information, CoT-Seg can augment the Reasoner with an external retrieval step. Specifically, a Retrieval Agent is invoked to search for relevant information from the web or a knowledge database, which is then incorporated into the chain-of-thought reasoning.

The Retrieval Agent searches for information about the person, such as reference images or textual descriptions, and provides these as additional inputs to the Reasoner. The Reasoner then integrates the retrieved knowledge into its CoT reasoning to generate a meta-query, e.g., specifying appearances, coloring, unique clothing, pose, or contextual cues, which guides the Segmentation Agent to correctly segment the target. This mechanism allows CoT-Seg to handle queries that require external or domain-specific knowledge, extending its reasoning capabilities beyond the information present in the original input.

%% file: sec/4_experiment.tex
\section{Experiments}\label{sec:experiments}
\subsection{Experimental Setup}
In our quantitative examples, we mainly focus on the ReasonSeg~\cite{lai2023lisa} dataset where reasoning segmentation is necessary compared to RefCOCO~\cite{kazemzadeh2014referitgame} where the queries are explicit and purposed for referring segmentation. However, ReasonSeg contains some imperfections, some prompts do not require deep reasoning (eg. "the tennis player"), while other prompts do not make sense (eg. "During a brainstorming event, it is common to record and present ideas on what object is the whiteboard in the room?").  

Given more recent and significant advancement in reasoning segmentation, a dataset update is due and necessary, which should contain more difficult cases such as closely connected objects and multiple objects similar to the object of interest, with implicit queries requiring complex reasoning to understand for challenging segmentation task. Thus, in this paper, we propose {\sc ReasonSeg-Hard}, a new evaluation dataset for stress testing reasoning segmentation. Specifically, we constructed a dataset with 213 image-query pairs consisting of 75 images and their respective queries sampled from ReasonSeg Test Split. We sample query-image pairs that either require deeper and more thorough reasoning to identify object(s) of interest, or queries including complex objects inherently difficult to segment due to size, transparency or surroundings. Refer to Appendix~\ref{app:reasonseg-hard} for data examples and additional details. We compare CoT-Seg against state-of-the-art reasoning segmentation methods including LISA~\citep{lai2023lisa}, GSVA~\citep{xia2024gsva}, Vision Reasoner~\citep{liu2025visionreasoner}.

As our method is training-free, we emphasize zero-shot evaluation to highlight the effectiveness of inference-time reasoning and self-correction. Performance is measured by Generalized Intersection-over-Union (gIoU) and Complete Intersection-over-Union (cIoU).

\subsection{Implementation Details}
For the reasoner and valuator modules, we use GPT-4o~\citep{hurst2024gpt4o} and  Vision-Reasoner-7B~\citep{liu2025visionreasoner} as the segmentation agent unless otherwise stated, with system prompts tailored for CoT reasoning, summarization, and self-correction. 
The chain-of-thought reasoning length is adaptively determined by the Reasoner, typically converging within 4--8 steps. 
The Segmentation Agent is instantiated with Vision-Reasoner-7B~\citep{liu2025visionreasoner} with SAM-HQ~\citep{ke2023segment} by default, though we also test compatibility with other SAM-based variants~\citep{kirillov2023segment}. 
Structured communication between Reasoner, Evaluator, and Segmentation Agent is implemented in JSON format to handle multimodal control inputs and to ensure capability alignment.

For cases that need further domain information, the user can enable retrieval-augmented reasoning. We employ a lightweight agent that queries the web using entity names or context keywords extracted by the Reasoner. Retrieved data is passed back as either text descriptions or reference images to the MLLM agent during the CoT process. This experiment shows CoT's ability to incorporate RAG, which leverages the capabilities of MLLM for segmentation purposes and shows the potential in vision-language integration. 
To ensure reproducibility, most experiments are ran on an NVIDIA 4090 GPU with 24GB memory, although the majority of reasoning computation using GPT-4o occurs in the cloud-hosted LLM. (Qwen and Gemma 3 experiments were ran on 2 and 3 NVIDIA 4090 GPUs. 

\vspace{-0.05in}
\subsection{Qualitative Evaluation}
\begin{figure*}[t]
    \centering
    \includegraphics[width=0.6\linewidth]{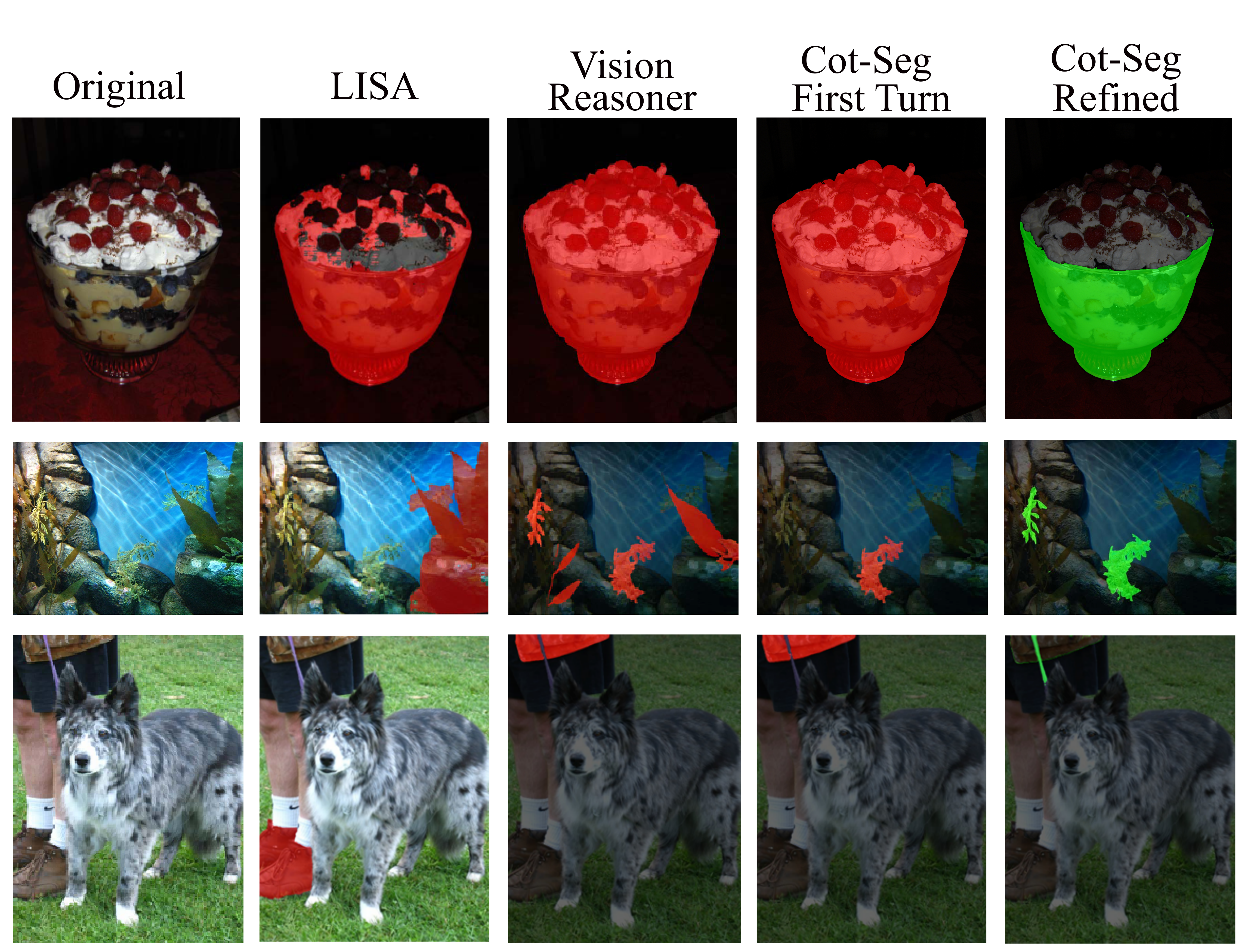}
    \vspace{-0.15in}
    \caption{Queries for each row are:
    1. A fruit salad is a refreshing and delicious
 dessert that often consists of a variety of fruits
 mixed together. What object in the picture could
 be used to hold and serve such a dessert?
 2. Please segment leafy sea dragons
 in this image.\\
 3. What is the object that the person in the
 picture is holding onto while walking his dog?}
\label{fig:qualitative_0}

\end{figure*}

\begin{figure*}[h]
\includegraphics[width =\linewidth]{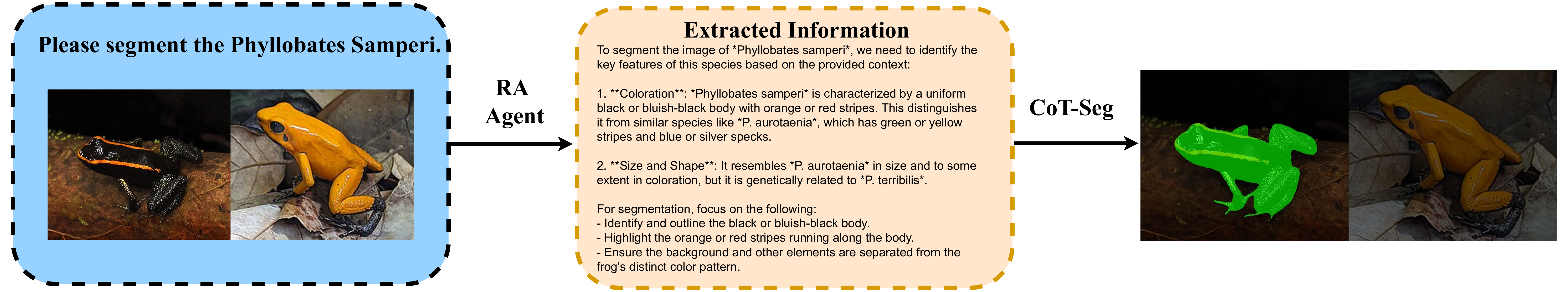}
\vspace{-0.2in}
\caption{
A recently discovered species of frog  unrecognizable to GPT-4o. With retrieval augmented (RA) reasoning CoT-Seg was able to segment the frog based on its appearance descriptions from the retrieval agent.
}
\vspace{-0.15in}
\label{fig:rag_frog}
\end{figure*}


We presented earlier qualitative  comparisons in Figures~\ref{fig:violinist}--\ref{fig:unracked}. More results in Figures~\ref{fig:qualitative_0}--\ref{fig:rag_frog} demonstrate how CoT-Seg progressively reasons about challenging queries and refines initial segmentation masks, demonstrating CoT-Seg's unique capabilities in:
1) resolving implicit queries with multi-step reasoning;
2) correcting masks with fine-grained self-correction (e.g., In Figure~\ref{fig:qualitative_0}, removing false positives such as ice cream in row 1 and recovering missed objects in row 2); and 3) retrieval-augmented reasoning for segmenting uncommon entities, such as identifying a new animal species (Figure~\ref{fig:rag_frog}) 
by integrating retrieved textual and visual cues. 
These results show that CoT-Seg achieves higher robustness in complex reasoning cases compared to prior methods that rely solely on direct prompt-to-mask predictions.
\subsection{Quantitative Evaluation}
\label{sec:quan}
\input{tab/referring}
We focus on reasoning segmentation tasks for our analysis: Tables~\ref{tab:reasonseghard} and~\ref{tab:reasonseg} summarize quantitative comparisons across benchmarks.
 CoT-Seg achieves SOTA or competitive results in both benchmarks, with the most improvements on {\sc ReasonSeg-Hard}, where high-level reasoning and domain knowledge are essential, while producing improved results after self-correction. 



\subsection{Ablation Studies}

\noindent \textbf{Impact of Self-Correction and CoT~~} Table~\ref{tab:reasonseg} and~\ref{tab:cot} compare performance with and without the refinement module, showing how  refinement improves robustness in ambiguous or cluttered scenes. Qualitative examples are shown in Figure~\ref{fig:qualitative_0}. Through refinement, CoT-Seg is capable of correcting for the missed object in the first turn results. We discovered that most of the time one round of auto-correction is enough and further correction rounds have not much effect.  

\input{tab/cot} Additionally, Table~\ref{tab:cot} studies the effect of CoT on segmentation. For segmentation without CoT, we modified our prompt for the MLLM so that the MLLM directly outputs the query naming the object to segment instead of going through the CoT query and answer process. The result shows that CoT improves the segmentation accuracy.

\input{tab/refcoco}
\vspace{0.05in}
\noindent \textbf{Easier General Benchmark~~} We examine  CoT-Seg with RefCOCO as an easier benchmark in Table~\ref{tab:refer}. In comparison with {\it ReasonSeg} and {\sc ReasonSeg-Hard}, where CoT-Seg significantly outperforms existing baselines, the improvement in RefCOCO is rather marginal. We attribute this result to the difficulty of benchmarks, where fewer examples in RefCOCO require long thinking process. In contrast, we achieve more significant improvements when processing more challenging data.

\noindent \textbf{Effect of Chain-of-Thought Length~~} We vary the number of reasoning steps (e.g., 2, 4, 8) to study the tradeoff between reasoning depth and segmentation quality. \input{tab/cotlength}
Table~\ref{tab:cotlength} tabulates the results where all of the experiments use a maximum of two rounds of refinements for self-correction running on {\sc ReasonSeg-Hard}. 
The results show that the length of chain of thoughts is not critical to performance, 
with a length of 4 producing the best score among the tested fixed lengths. 
Fixed CoT length is outperformed by variational length(averaging gIoU and cIoU)
determined by the MLLM. The results indicates that two reasoning steps usually suffice while overthinking with too many  steps may  lower the accuracy, with varying lengths depending on the input results in the best accuracy. 

\input{tab/compatibility}
\noindent \textbf{Segmentor Compatibility~~} In our quantitative experiments, the Segmentation Agent can use different segmentation backbones. We analyze how their  capabilities  affect downstream performance. Table~\ref{tab:cotcompatibility} tabulates the results, highlighting the importance of segmentor capability descriptions in guiding Reasoner and Segmentator collaboration. 

\noindent \textbf{MLLM Agent Variants~~} 
We evaluate CoT-Seg with different MLLM backbones, such as GPT-4o, Gemma 3 12b, and Qwen2.5-VL-7B on {\sc ReasonSeg-Hard} with maximum of 2 rounds of refinement. Table~\ref{tab:cotmllm} tabulates the results, which  reveal how reasoning depth, hallucination tendency, and multimodal grounding influence segmentation quality and stability, showing the tradeoffs between proprietary and open-source models in reasoning-driven segmentation. For earlier VL models such as Qwen2.5, when given two segmentations, they cannot determine which one is better so they can only fulfill the CoT part and not the auto-correction part of our framework.
\input{tab/mllmcompare}

\begin{figure}
\centering
\includegraphics[width=\linewidth]{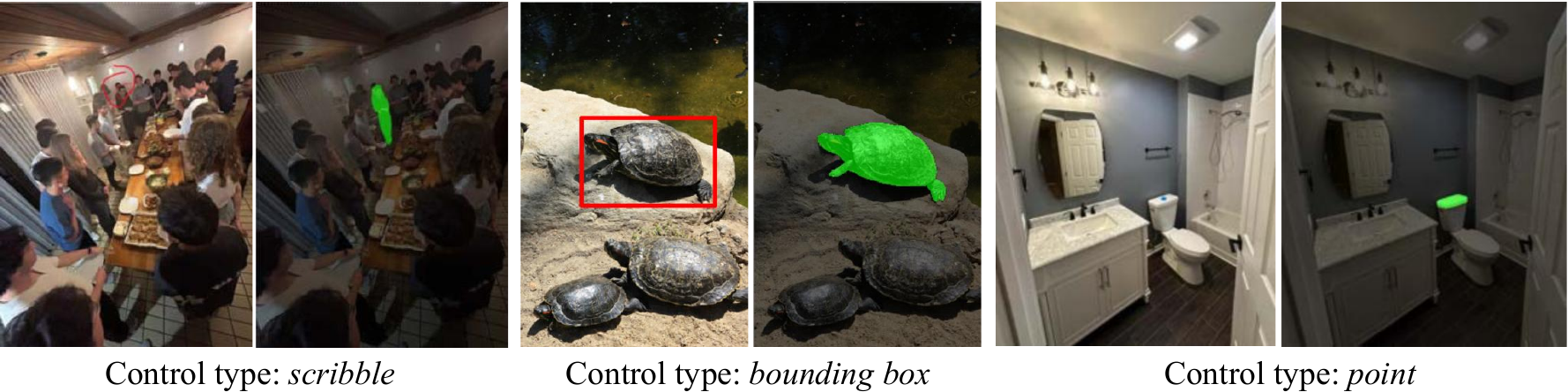}
\vspace{-0.15in}
\caption{Multimodal inputs. CoT-Seg supports diverse control types, such as scribble, bounding box, and point, allowing users to easily interact with.}
\label{scribble}
\vspace{-0.1in}
\end{figure}

\noindent \textbf{Multimodal Input Control~~}
Our framework can be used for multiple kinds of input including but not limited to bounding box, point, and scribble annotations, demonstrating the flexibility of JSON-based multimodal reasoning and how CoT and auto-correction works for all general reasoning strategies, shown in Figure~\ref{scribble}. CoT works especially well on improving segmentation based on rough human input, providing important text info for the segmentation agent.

\input{tab/time}
\noindent \textbf{Limitation: Inference Overhead~~} Table~\ref{tab:time} shows the inference overhead of CoT-Seg compared to other models. CoT-Seg (Avg) is the average inference time on the {\sc ReasonSeg-Hard} dataset (some image-query pairs require auto correction while others do not). In our quantitative experiments, CoT-Seg has significant overhead due to GPT online API calls, trading time for higher accuracy.    



%% file: tab/referring.tex
\begin{table}

\begin{minipage}[t]{0.5\textwidth}
    \centering
\caption{Reasoning segmentation evaluation with complex and implicit queries on our {\sc ReasonSeg-Hard} dataset.  \textdagger~is produced with the official released weights with 8-bit quantization.}
\setlength{\tabcolsep}{5.5mm}{
\renewcommand\arraystretch{1.0}
\resizebox{0.85\linewidth}{!}{%
\begin{tabular}{l|ccc}
\toprule
Method & gIoU & cIoU \\
& & \\
\midrule
\midrule
LISA-13B-Llama2\textsuperscript{\textdagger} ~\citep{lai2023lisa} & 38.0 & 41.1 \\
GSVA-7B~\citep{xia2024gsva}  &40.9&  37.8 \\
Vision-Reasoner-7B~\citep{liu2025visionreasoner} & 49.1 & 48.1 \\ 
SegZero-7B~\citep{liu2025seg} &44.0&52.6 
 
\\\midrule
CoT-Seg &\underline{56.7}&\underline{54.4}\\
CoT-Seg with self-correction & \bf{58.6}&\bf{57.4}\\
\bottomrule 
\end{tabular}
\vspace{0.1in}
\label{tab:reasonseghard}
}
}
\end{minipage}
\hfill
\begin{minipage}[t]{0.5\textwidth}
    \centering
\caption{Quantitative evaluation on the test set of \textit{ReasonSeg}~\citep{lai2023lisa}. (ft) means finetuning on the train set. \textdagger~is reproduced with the official released weights with 8-bit quantization.}
\setlength{\tabcolsep}{5.5mm}{
\renewcommand\arraystretch{1.0}
\resizebox{0.85\linewidth}{!}{%
\begin{tabular}{l|c c}
\toprule
Method&\multicolumn{2}{c}{ReasonSeg (overall)}\\
&gIoU&cIoU\\\midrule\midrule
OVSeg~\citep{liang2023open}&26.1&20.1\\
GRES~\citep{liu2023gres}&21.3&22.0\\
X-Decoder~\citep{zou2023generalized}&21.7&16.3\\
SEEM~\citep{zou2023segment}& 24.3&18.7\\
Seg-Zero-7B~\citep{liu2025seg}&57.5&52.0\\
Vision-Reasoner-7B~\citep{liu2025visionreasoner}& 63.6&-\\
LISA-13B~\citep{lai2023lisa} &44.8&45.8\\
LISA-13B-Llama2 (ft)\textsuperscript{\textdagger}~\citep{lai2023lisa}&{50.0}&51.9\\
LISA-13B-LLaVA1.5 (ft)~\cite{lai2023lisa}&61.3&\bf{62.2}\\
\midrule
CoT-Seg&\underline{66.0}&58.8\\
CoT-Seg with self correction &\bf {66.7}&\underline{60.4}\\
\midrule
\end{tabular}
\label{tab:reasonseg}
}
}
\end{minipage}
\vspace{-0.3in}

\end{table}

%% file: tab/cot.tex
\begin{table}
\centering
\caption{Ablation study of CoT on {\sc ReasonSeg-Hard}} 
\vspace{-0.10in}
\resizebox{0.8\columnwidth}{!}{%

\renewcommand\arraystretch{1.0}

\begin{tabular}{l|c c}
\toprule
\multirow{2}*{CoT Length}&\multicolumn{2}{c}{ReasonSeg-Hard}\\
&gIoU&cIoU\\\midrule\midrule
Seg without CoT &51.5&53.5\\
CoT-Seg with no auto-correction & 56.7 & 54.4 \\
CoT-Seg with 1 round of auto-correction  &\textbf{58.6}&\textbf{57.4}\\
CoT-Seg with 2 round of auto-correction  &58.4& 57.3\\
\bottomrule     
\end{tabular}
\label{tab:cot}
}
\vspace{-0.2in}
\end{table}

%% file: tab/refcoco.tex
\begin{table}[t]
\centering
\caption{Referring expression segmentation results on RefCOCO~\citep{kazemzadeh2014referitgame}. 
The cIoU metrics for each split are reported. 
}
\resizebox{.85\linewidth}{!}{

\begin{tabular}{l|ccc}
\toprule
Method & Val. & Test-A & Test-B \\
\midrule
\midrule
MCN~\citep{luo2020multi} & 62.4 & 64.2 & 59.7 \\
VLT~\citep{ding2021vision} & 67.5 & 70.5 & 65.2 \\
CRIS~\citep{wang2022cris} & 70.5 & 73.2 & 66.1 \\
LAVT~\citep{yang2022lavt} & 72.7 & 75.8 & 68.8 \\
LISA-7B (fine-tuned on ReferSeg)~\citep{lai2024lisa} & 74.9 & 79.1 & 72.3 \\
Seg-Zero 7B~\citep{liu2025seg} & -- & 80.3 & -- \\
GSVA-7B (ft)~\citep{xia2024gsva} & \textbf{77.2} & 78.9 & 73.5 \\
\midrule
CoT-Seg & \underline{76.3} & \underline{80.9} & \underline{72.7} \\
CoT-Seg with self-correction & \textbf{77.2} & \textbf{80.9} & \textbf{73.8} \\
\bottomrule
\end{tabular}
}

\label{tab:refer}
\vspace{-0.05in}
\end{table}

%% file: tab/cotlength.tex
\begin{table}
\centering
\caption{CoT length experiment on \sc ReasonSeg-Hard}
\vspace{-0.10in}
\setlength{\tabcolsep}{5.5mm}{
\renewcommand\arraystretch{1.0}
\resizebox{\linewidth}{!}{%
\begin{tabular}{l|c c}
\toprule
\multirow{2}*{CoT Length}&\multicolumn{2}{c}{ReasonSeg-Hard}\\
&gIoU&cIoU\\\midrule\midrule
CoT-Seg with self-correction using CoT length 2&53.9&48.6\\
CoT-Seg with self-correction using CoT length 4&50.4&\textbf{58.8}\\
CoT-Seg with self-correction using CoT length 8&52.9&51.7\\
CoT-Seg with self-correction using CoT variational length &\textbf{58.6}&57.4\\
\bottomrule
\end{tabular}
\label{tab:cotlength}
}
}
\end{table}

%% file: tab/compatibility.tex
\begin{table}
\centering
\caption{Segmentor experiment without self-correction {\sc ReasonSeg-Hard}}
\vspace{-0.1in}
\setlength{\tabcolsep}{5.5mm}{
\renewcommand\arraystretch{1.0}
\resizebox{\linewidth}{!}{%
\begin{tabular}{l|c c}
\toprule
\multirow{2}*{Segmentor}&\multicolumn{2}{c}{ReasonSeg-Hard}\\
&gIoU&cIoU\\\midrule\midrule
CoT-Seg with Vision-Reasoner-7B + SAM-HQ&\bf{56.7}&54.4\\
CoT-Seg with LISA &45.3&49.7\\
CoT-Seg with SegZero-7B &44.1&\bf{56.4}\\
\bottomrule
\end{tabular}
\label{tab:cotcompatibility}
}
}
\end{table}

%% file: tab/mllmcompare.tex
\begin{table}
\centering
\caption{Different MLLM experiments on \sc{ReasonSeg-Hard}}
\vspace{-0.1in}
\setlength{\tabcolsep}{5.5mm}{
\renewcommand\arraystretch{1.0}
\resizebox{\linewidth}{!}{%
\begin{tabular}{l|c c}
\toprule
\multirow{2}*{CoT with different MLLMs}&\multicolumn{2}{c}{ReasonSeg-Hard}\\
&gIoU&cIoU\\\midrule\midrule
CoT-Seg with self-correction using GPT-4o&\bf{58.6}&\bf{57.4}\\
CoT-Seg with self-correction using Gemma 3-12B &49.8&54.5\\
CoT-Seg with self-correction using Qwen2.5-VL-7B&42.4&51.2\\
\bottomrule
\end{tabular}   
\label{tab:cotmllm}
}
}
\end{table}

%% file: tab/time.tex
\begin{table}
\centering
\caption{Inference time (In seconds)} 
\vspace{-0.15in}
\setlength{\tabcolsep}{5.5mm}{
\renewcommand\arraystretch{1.0}
\resizebox{0.8\linewidth}{!}{%
\begin{tabular}{l|c c}
\toprule
\multirow{2}*{CoT Length}&\multicolumn{2}{c}{ReasonSeg-Hard}\\
&1 Image-Query Pair\\\midrule\midrule
LISA-13B~\citep{lai2023lisa} &1.26\\
GSVA &3.03\\
Vision Reasoner~\citep{liu2025visionreasoner}&4.31\\

CoT-Seg without need of self correction&37.846\\
CoT-Seg with 2 rounds of self correction&83.386\\
CoT-Seg (Avg)&67.077\\

\bottomrule 
\end{tabular}
\label{tab:time}
}
\vspace{-0.2in}
}
\end{table}

%% file: sec/5_conclusion.tex
\section{Conclusion}
We introduced \textbf{CoT-Seg}, a zero-shot framework that rethinks reasoning segmentation by integrating chain-of-thought reasoning and self-correction with off-the-shelf MLLMs and segmentation agents. Our method enables step-by-step reasoning to synthesize meta-queries, collaborative evaluation for refinement, and retrieval-augmented reasoning for knowledge gaps. We have also proposed a new dataset ReasonSeg-Hard to test the effects of CoT on difficult scenarios. This work highlights the untapped potential of inference-time reasoning and self-correction in bridging vision-language understanding with precise segmentation.

%% file: sec/6_Appendix.tex
\appendix
\renewcommand{\thesection}{Appendix \Alph{section}}
\section{MLLM Prompt Details}

\paragraph{CoT First Turn Template\\}

We use this as a basic description for the LLM to propose questions answer pairs for CoT process, we replace \texttt{<}\textit{QUERY}\texttt{>} with the user query.\\
\textit{
You will serve as an agent for language-based image segmentation model. During each inference, your task is to consider a query and describe a given image with chain of thoughts. You need to provide details to help the segmentation model understand the image better. The target objects may contain multiple layers, be blocked by other object, or be seamlessly embedded in their surroundings. Your description will be later sent to the segmentation as prompt. For example, if given an image, you need to describe what can be seen in the image, the number of objects for each categories, the position of the target object, the structure of the object, the number of layers of the object, etc. The actual description depends on the given image.For the output, you need to follow the format:- Question 1: Answer 1.- Question 2: Answer 2 ..., etc, where each pair of prompt and answer implies the chain of thoughts, i.e., different levels or different part of the image understanding. For example, the first prompt can be related to the overall style or background of the image. Finally, you need to summarize the description based on your generated prompts and answers with strictly with the format: Your summary here ... considering the prompt where the user is looking for ..., the object of interest may be ... Then, based on the summary, you have to generate a pseudo-prompt to query the segmentation model. This pseudo-prompt should contains the information about what is in the image, what to segment, and where the target object is. It must strictly follow the format: - Prompt: The image shows ....Please segment the ... located at ... of the image. Lastly, please generate a list of labels that would be passed to an object detector based on the summary and thinking process that describes the object(s) that fits the user query and should be segmented and in strictly in the format of cat. remote control. television. with a period separating every label and if there is multiple word in a label then separate using space. Do not include other objects not given by the prompt.
} 
\texttt{<}\textit{QUERY}\texttt{>}

\paragraph{CoT Self-Correction Template\\}
This is the template to extract meta-queries for self-correction if needed.

\textit{
You will serve as an agent for language-based image segmentation model. You need to decide whether the segmentation result is good or not. If it is not good, you need to provide the meta-queries for refinement. During each inference, you will be given a pair of images and a user query, one of which is the original image, and the other one with blank background is the segmentation result with respect to the query after masking. Your task is to describe the pair of given images with chain of thoughts and decide whether the segmentation result correctly reflects the user query. The segmentation result should include all the objects related to the user query, and should not contain any other objects or distinctions unrelated to the user query; otherwise, it will be considered incorrect. A correct segmentation result is expected to contain objects of interest isolated in a white background, the segmentation result can be fragmented if some parts of the object is obscured by obstacles. You may ignore small artifacts/noises in the background. }

\textit{
For the output, you need to follow the format:
- Reasoning process: 
1. Original image: $<$ reasoning on the original image$>$.
2. Segmentation image: $<$reasoning on the segmentation image$>$.
3. Summary: $<$reasoning on the correctness of the segmentation$>$,
- Correctness: $<$correctness$>$True$<$/correctness$>$
- Meta-queries (Output if the correctness is false):
1. Positive: $<$positive$>$None or Please also segment the xxx, located at ... of the original image.$<$/positive$>$
2. Negative: $<$negative$>$None or Please remove the xxx, located at ... of the segmentation image.$<$/negative$>$
- Labels:
1. Positive: $<$plabels$>$label1. label2. label3.$<$/plabels$>$
2. Negative: $<$nlabels$>$labels.$<$/nlabels$>$
}
\textit{
Specifically, during the reasoning, you have to decide what the query refers to, what can be seen in the image, where the target object is, how many target object is, and more. The correctness is True if no refinement on the segmentation image is needed. If correctness is False, please output a positive and a negative meta-query. The positive meta-query implies what needs to be added (false negative), and the negative meta-query implies refers to what needs to be removed (false positive). The positive meta-query is 'None' if the segmentation has already included all the target object, i.e. no false negatives and you don't want to include anything additional. Similarly, the negative meta-query is 'None' if you do not want to remove things from the segmentation. In the negative meta-query, your description should be solely based on the segmentation image, i.e., do not use reference to the original image (for example, next to something that you cannot observe in the segmentation image). Your meta-queries will be sent to the language-based segmentation model for refinement, so please keep your meta-queries clear and understandable following the format. Finally, please for the corresponding positive meta query and negative meta query please also output labels that corresponds to the object(s) being added or removed, period separated and if there is more than one word in a label separate using whitespace. Do not include surrounding object or background of target object in positive meta-query. 
}

\section{MLLM Sample Outputs}
\begin{figure}[h]
\centering  
\includegraphics[width =.85\linewidth]{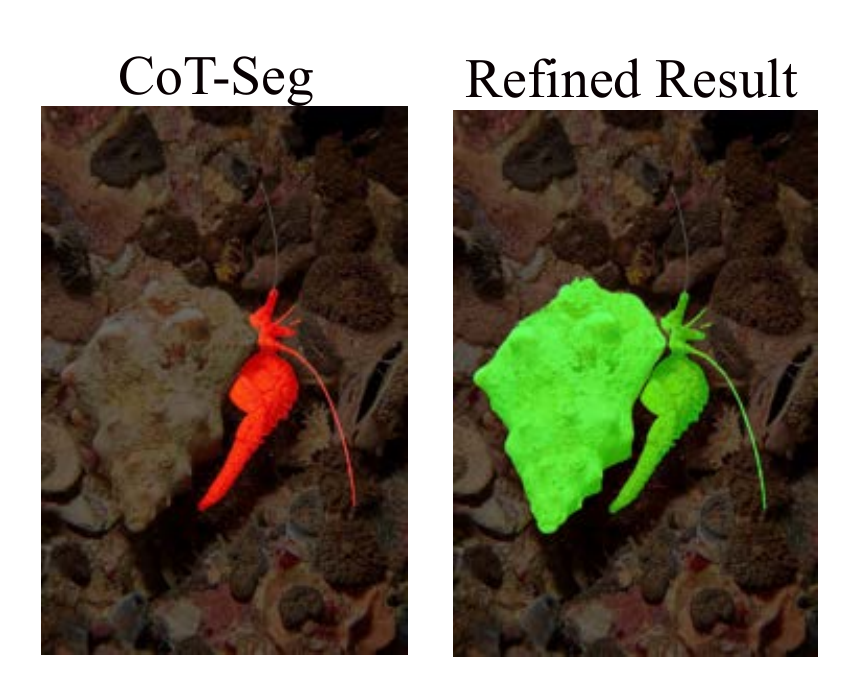}
\caption{Prompt: Segment the Pagurian}
\label{fig:pagurian}
\end{figure}

\paragraph{CoT Self-Correction Process For Figure~\ref{fig:pagurian}\\}

\textit{- Reasoning process:}
\\
\textit{1. Original image: The original image shows an underwater scene with various marine life forms and corals. The Pagurian, or hermit crab, is located at the center-right of the image. It is identifiable by its orange color and the shell it occupies.}
\\
\textit{
2. Segmentation image: The
segmentation image shows an
isolated orange Pagurian on a white
background. The shell is not
included in the segmentation.
}
\\
\textit{
3. Summary: The segmentation
correctly isolates the Pagurian itself
but fails to include the shell, which is
an integral part of the Pagurian's
appearance in the original image.
Therefore, the segmentation is
incomplete.
}
\\
\textit{
- Correctness: False}
\\
\textit{
- Meta-queries (Output if the
correctness is false):
}
\\
\textit{
Positive: Please also
segment the shell of the Pagurian,
located at the center-right of the
original image.
}
\\
\textit{
Negative: None
}
\\
\textit{
- Labels:
}
\\
\textit{
Positive: shell
}
\\
\textit{
Negative: None.}\\

\begin{figure}[h]
\centering  
\includegraphics[width =.85\linewidth]{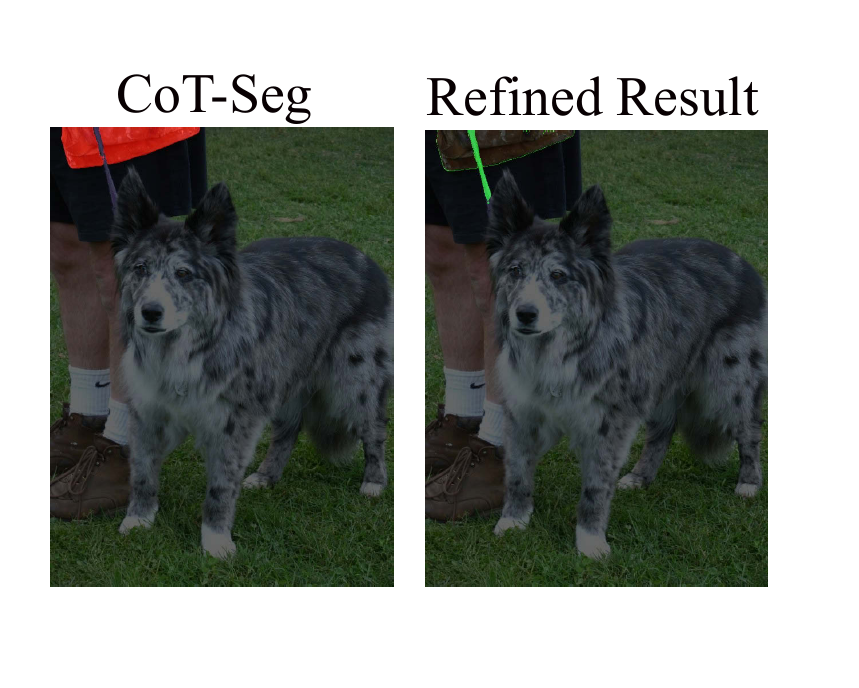}
\caption{Prompt: What is the object that the person in the picture is holding onto while walking his dog?}
\label{fig:dog_supp}
\end{figure}

\paragraph{Full First Turn and Self-Correction CoT for Figure~\ref{fig:dog_supp}}

\textit{
\\
\textbf{First Turn}\\
- Question 1: What is the overall setting of the image?
\\
- Answer 1: The image shows a person standing on grass, with a dog in the foreground.
\\
- Question 2: What are the main objects visible in the image?
\\
- Answer 2: The main objects visible are a person, a dog, and a leash.
\\
- Question 3: What is the person holding?
\\
- Answer 3: The person is holding a leash.
\\
- Question 4: Where is the object of interest located in the image?
\\
- Answer 4: The leash is located in the upper left part of the image, extending from the person's hand to the dog.
\\
Summary: The image shows a person standing on grass with a dog. Considering the prompt where the user is looking for the object that the person is holding while walking their dog, the object of interest may be the leash.
\\
Prompt: The image shows a person standing on grass with a dog. Please segment the leash located at the upper left part of the image.
\\
\textbf{Self-Correction}\\
- Reasoning process: 
\\
1. Original image: The original image shows a person standing on grass with a dog. There is a leash visible in the upper left part of the image, held by the person.
\\
2. Segmentation image: The segmentation image shows a portion of the person's clothing and a small part of the leash in the upper right corner, isolated on a white background.
\\
3. Summary: The segmentation result does not correctly reflect the user query. The query asks for the leash to be segmented, but the segmentation image only includes a small part of it and part of the person's clothing. The entire leash should be included, and the person's clothing should be excluded.
\\
- Correctness: False
\\
- Meta-queries (Output if the correctness is false):
\\
1. Positive: Please also segment the entire leash, located at the upper left part of the original image.
\\
2. Negative: Please remove the person's clothing, located at the upper right part of the segmentation image.
\\
- Labels:
\\
1. Positive: leash.
\\
2. Negative: person's clothing.
}

\section{Additional Results}

\paragraph{Benchmarks on COD10K}
Results in Table~\ref{tab:cod} show that CoT-Seg displays SOTA or competitive results regarding segmenting highly difficult camouflaged objects, where CoT reasoning will make the task easier.
\input{tab/cod10k}

\begin{figure*}[h]
\centering
\includegraphics[width=\linewidth]{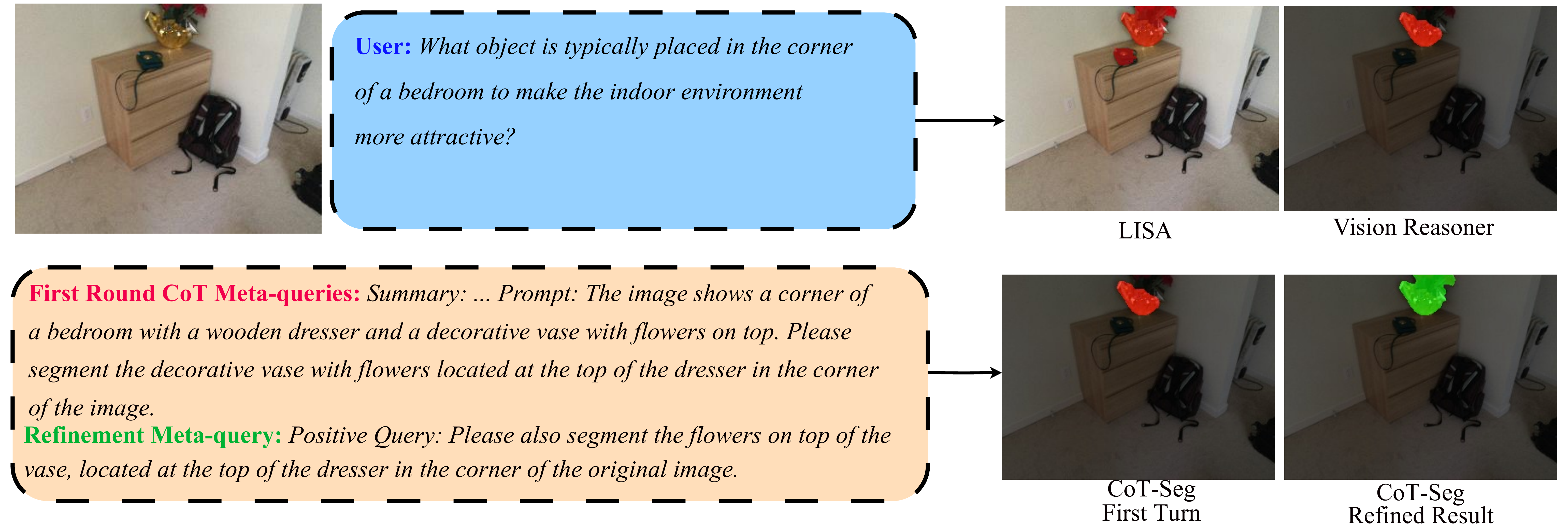}
\caption{CoT-Seg can correct minor mistakes such as not segmenting the flowers with the vase.}
\label{fig:qualitative_4}
\end{figure*}

\paragraph{Self-Correction Visual Examples}We show additional self-correction examples in Figure~\ref{fig:qualitative_4}.

\section{Similar Works Analysis and Comparison}
\paragraph{Vision Reasoner }
We discuss the difference between our work and the concurrent work VisionReasoner~\citep{liu2025visionreasoner}. To the best of our knowledge VisionReasoner uses reinforcement learning to generate the bounding boxes and segmentations. VisionReasoner has greatly improved on previous reasoning segmentation models as show in Tables~\ref{tab:reasonseghard} and~\ref{tab:refer} but still fails in some complicated cases where there are a large number of objects to be segmented Figures~\ref{fig:violinist}-\ref{fig:bicep} or when the prompt is very implicit. CoT-Seg in comparison, is zero-shot and can be easily plugged in to different models, offering high flexibility and achieves  higher scores in all the test data in Tables~\ref{tab:reasonseghard}, ~\ref{tab:reasonseg} and ~\ref{tab:refer}.

\paragraph{GSVA}
Table~\ref{tab:GSVA} shows CoT-Seg's competitive performance on the standard referring dataset RefCOCO with less emphasis on CoT deep reasoning for complex segmentation.
\begin{table}[h]
\centering
\begin{tabular}{l|cccc}
\toprule
Method& Val. & Test-A & Test-B \\
& & \\
\midrule
\midrule
GSVA-Llama2-13B~\citep{xia2024gsva} & \underline{77.7} & 79.9 & \underline{74.2}\\
GSVA-Llama2-13B (ft)~\citep{xia2024gsva} &\textbf{79.2} & \textbf{81.7} & \textbf{77.1}  
\\\midrule
CoT-Seg &76.3&80.9&72.7\\
CoT-Seg with self-correction & 77.2&\underline{80.9}&73.8\\
\bottomrule 
\end{tabular}
\caption{Quantitative comparison with
GSVA~\citep{xia2024gsva} on RefCOCO}
\label{tab:GSVA}
\end{table}

\begin{figure}[h]
\centering  
\includegraphics[width =\linewidth]{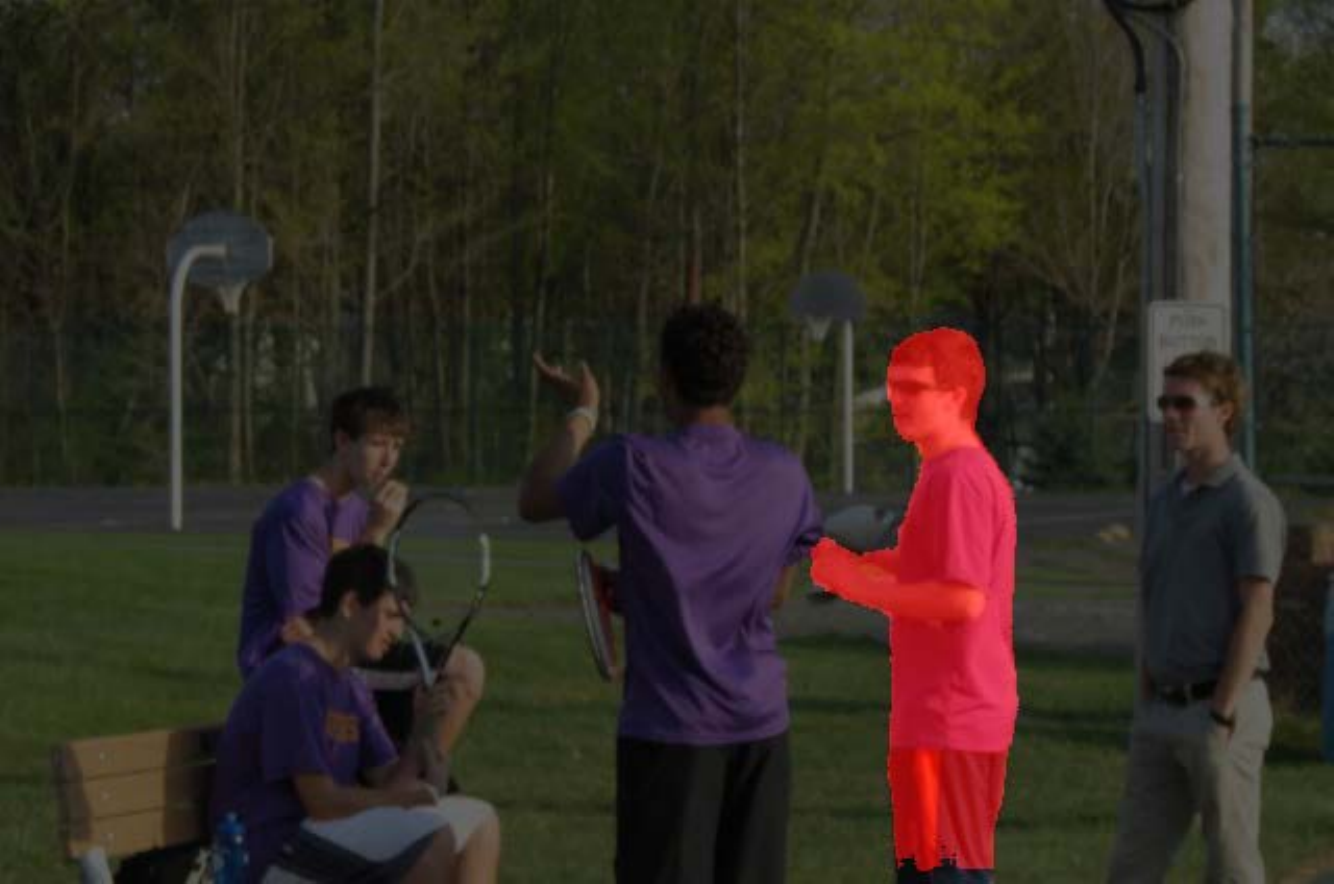}
\caption{GSVA - Prompt: Second from the right.}
\label{fig:teamgsva}
\end{figure}

\begin{figure*}[h]
\includegraphics[width =0.9\linewidth]{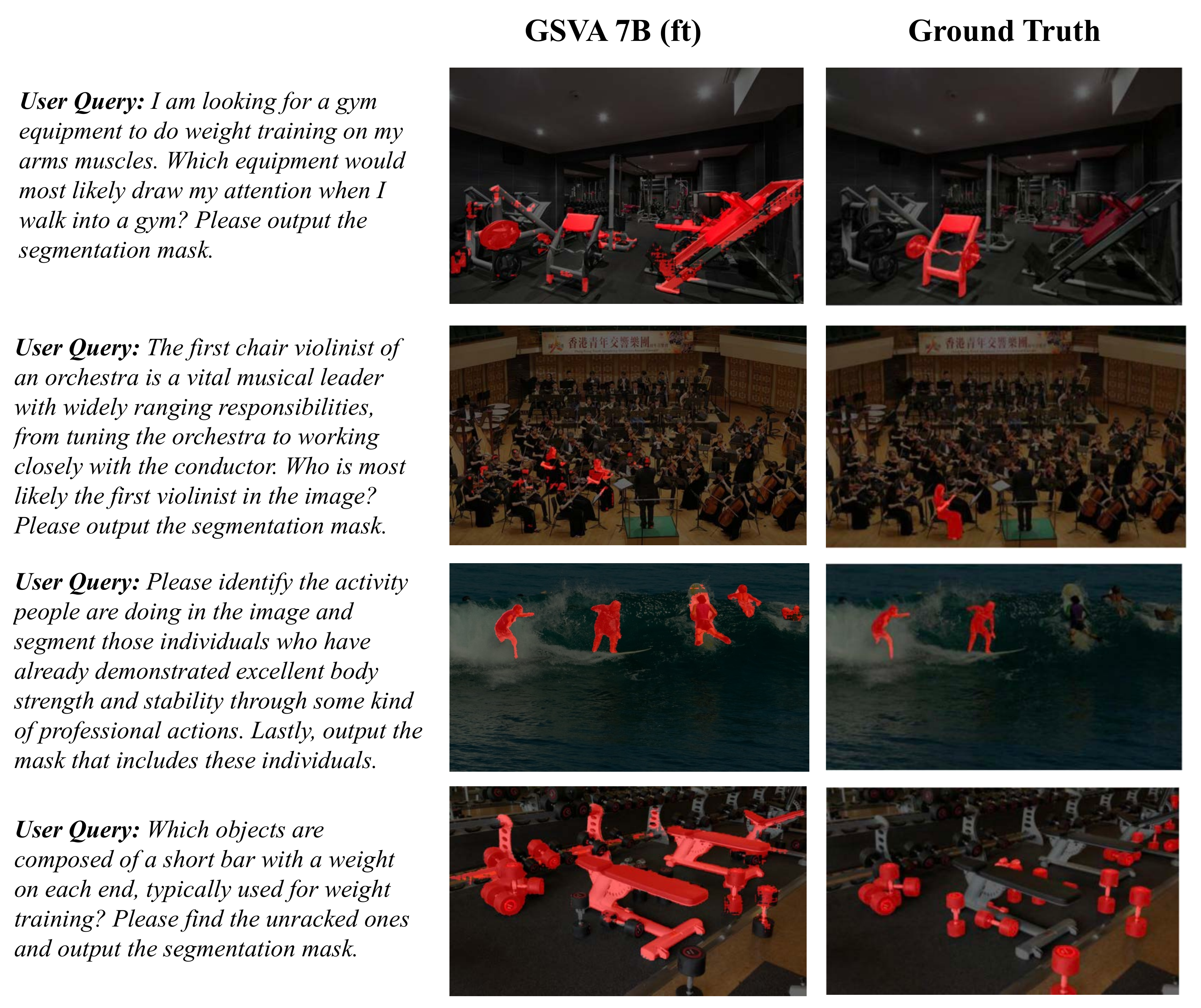}
\hspace{-1.15in}
\caption{GSVA inference results.}
\label{fig:gsva_examples}
\end{figure*}

GSVA~\citep{xia2024gsva} also uses MLLM to guide segmentation. Specifically, GSVA uses MLLM to generate [SEG] tokens and prompt the segmentation model to support multiple object segmentation and a [NULL] token to reject absent object. In comparison, our approach uses chain of thought reasoning to assimulate and provide useful information to the segmentator agent, empowering our model  to solve very implicit queries and achieve multiple-object segmentation in a training-free manner. Our auto-correction process further leverages MLLM to improve and obtain accurate segmentations that the segmentor agent cannot achieve on its own.  In RefCOCO tests in Table~\ref{tab:GSVA}, GSVA achieves slightly higher results, mainly because of training and finetuning on the RefCOCO training dataset getting higher accuracy in prompts containing numerical positional arguments. e.g.,  an example is shown in Figure~\ref{fig:teamgsva}. Usage of improved and finetuned models should be able to improve our results on these benchmarks, as well as incorporating our framework as a plugin to GSVA and other recent SOTA reasoning segmentation agents. On the other hand GSVA does not focus on reasoning segmentation and as a cIoU score of 43.4 for 7B model and 44.6 for 13B model on the Reasonseg dataset. Furthermore, inference on GSVA 7B (ft) shows that GSVA 7B (ft) was unable to get correct segmentation results when the prompt becomes more implicit such as Figures ~\ref{fig:violinist}--\ref{fig:unracked}, the inference results are shown in Figure~\ref{fig:gsva_examples}.



\section{ReasonSeg-Hard}\label{app:reasonseg-hard}
\begin{figure*}[h]
\centering  
\includegraphics[width =\linewidth]{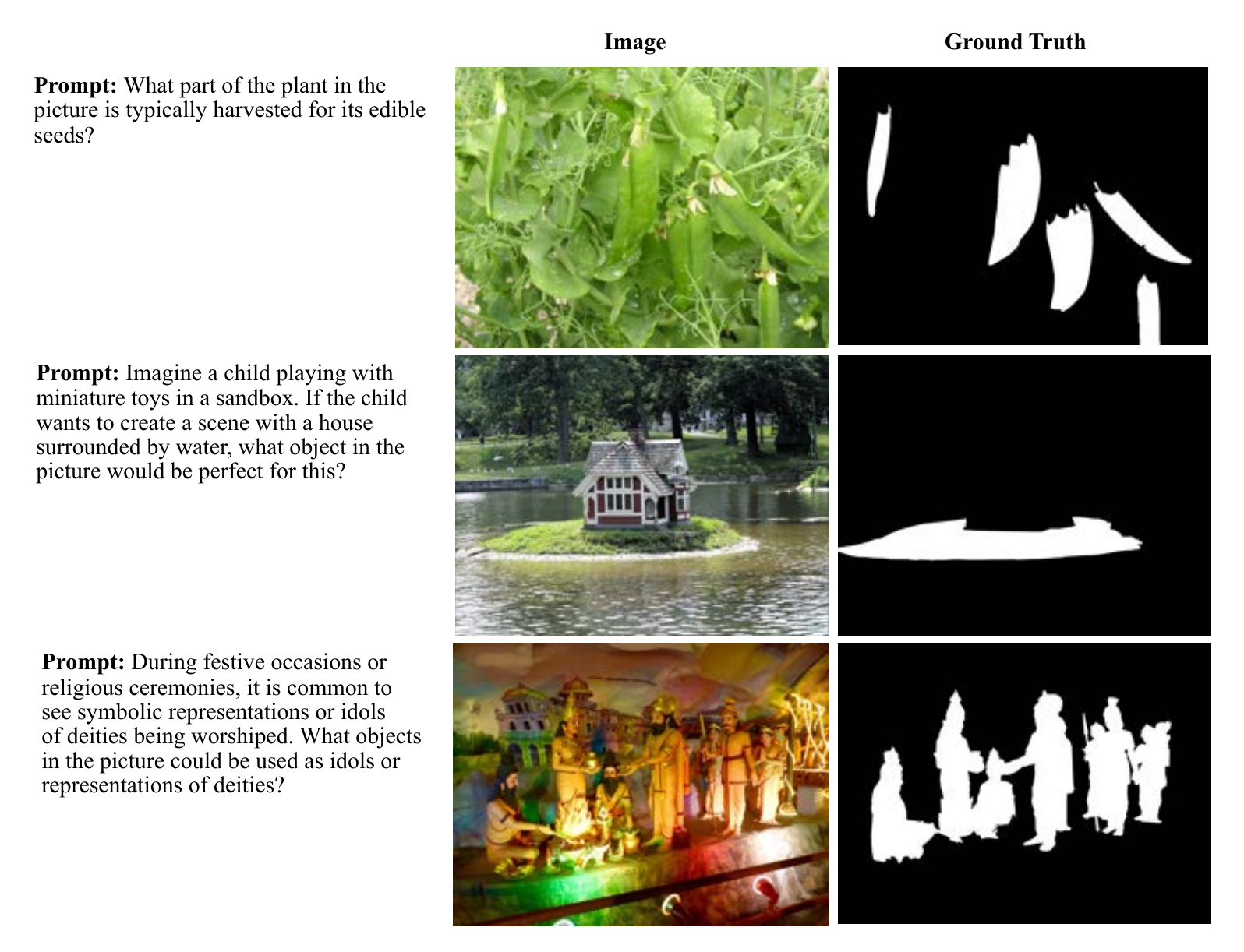}
\caption{Samples of challenging examples in {\sc ReasonSeg-Hard}.}
\label{fig:reasonsegds}
\end{figure*}

\begin{figure*}[h]
\centering  
\includegraphics[width=0.6\linewidth]{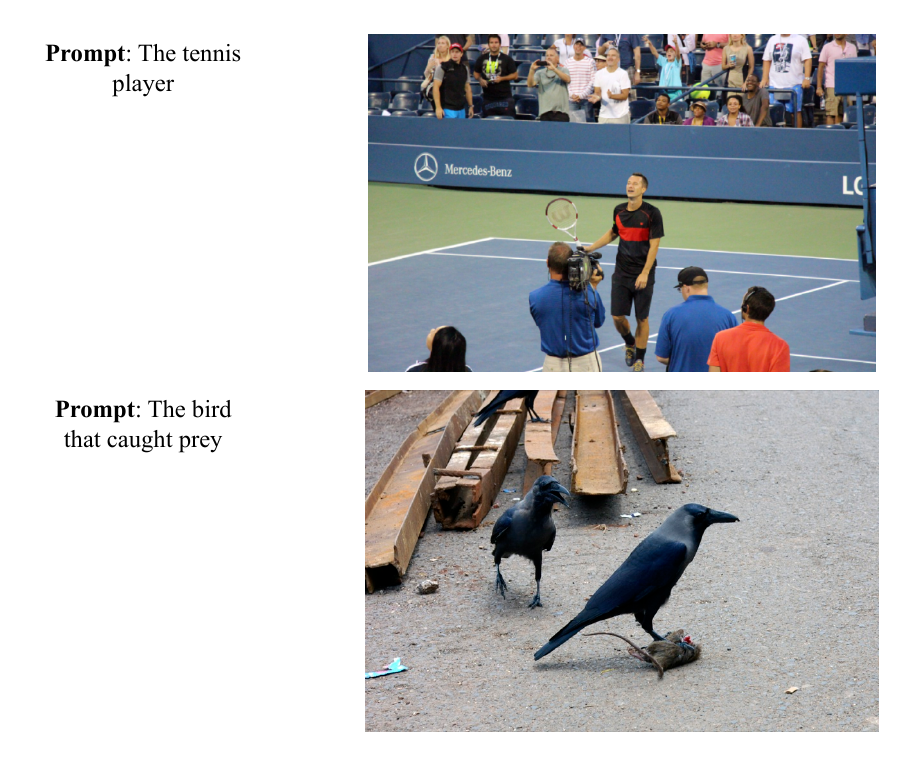}
\caption{Examples from ReasonSeg~\cite{lai2023lisa} which are excluded from {\sc ReasonSeg-Hard}  }
\label{fig:reasonsegex}
\end{figure*}

We sample implicit queries like ``\textit{When preparing for a festive event like Halloween, people often use certain objects to decorate their homes. What object in the picture would be suitable for this purpose?}" and excluded queries-image pairs that may be too simple such as ``\textit{something that the person uses to fish}". Examples     are shown in Figure~\ref{fig:reasonsegds}-\ref{fig:reasonsegex}.

\begin{figure*}[h]
\includegraphics[width =0.9\linewidth]{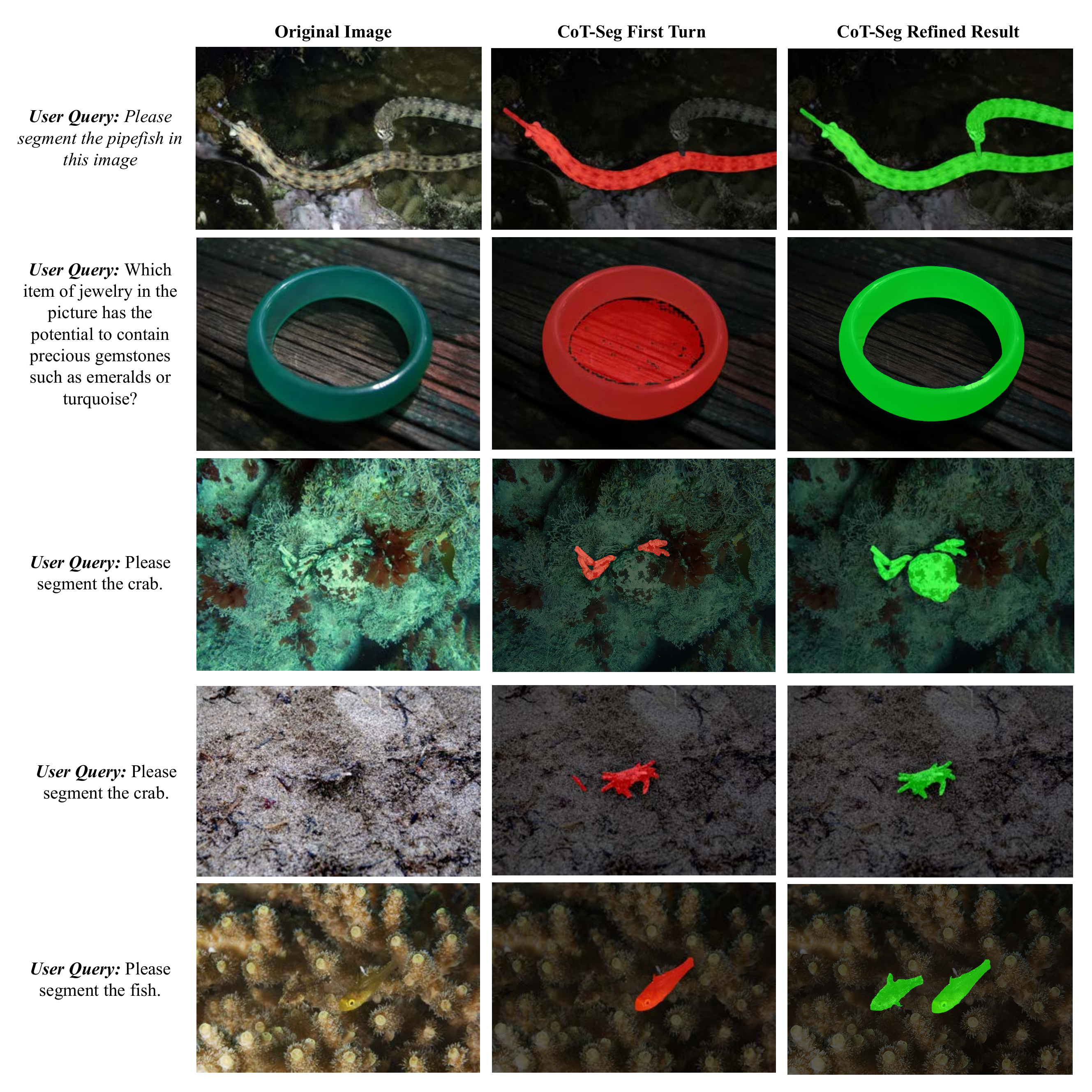}
\label{fig:refine_diagram}
\hspace{-1.15in}
\caption{Additional self-correction results.}
\end{figure*}

\begin{figure*}[h]
\includegraphics[width =\linewidth]{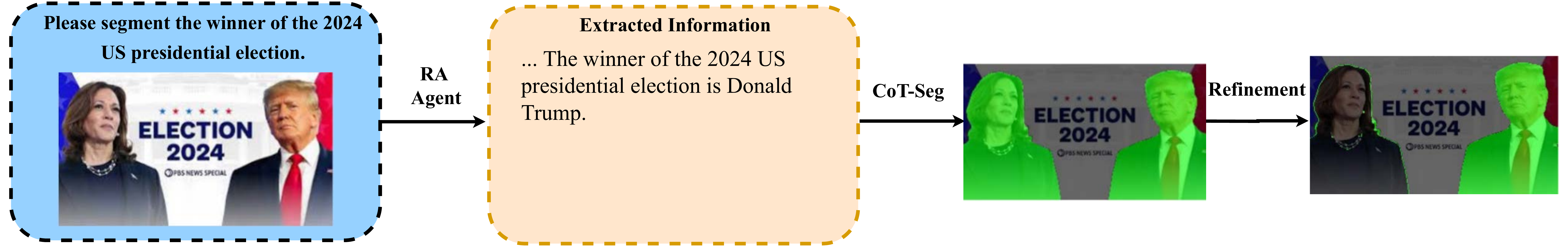}
\label{fig:election}

\caption{Retrieval-augmented CoT-Seg Result.}
\end{figure*}

%% file: tab/cod10k.tex
\begin{table}[h]
    \centering
    \caption{Quantitative evaluation on camouflaged images in \textit{COD-10K}~\citep{fan2020camouflaged}. Note that the models are blind to both dataset divisions. CoT-Seg with refinement improves on implicit query where with one round of segmentation may not be enough to fulfill the query.}
    \resizebox{\linewidth}{!}{%
    \begin{tabular}{l|c|cc|cc|cc|cc}
    \toprule
    
        &\multicolumn{1}{c|}{}&\multicolumn{4}{c|}{Train split}&\multicolumn{4}{c}{Test split}\\
        Method&\multicolumn{1}{c|}{Training free}&\multicolumn{2}{c|}{Implicit query}&\multicolumn{2}{c|}{Explicit query}&\multicolumn{2}{c|}{Implicit query}&\multicolumn{2}{c}{Explicit query}\\\cmidrule(lr){3-9}
         &&gIoU&cIoU&gIoU&cIoU&gIoU&cIoU&gIoU&cIoU\\ \midrule \midrule
         LISA-13B-Llama2~\citep{lai2024lisa}&No&65.0&60.3&66.5&59.7&63.4&55.8&66.5&58.0\\
         Vision-Reasoner-7B~\citep{liu2025visionreasoner}&No&77.1&77.1&\bf{77.6}&\underline{76.6}&76.9&76.3&77.4&{\bf 75.0}
         \\\midrule
         CoT-Seg&Yes&\underline{77.9}&{\bf 78.8}&\underline{77.5}&\bf{78.0}&\underline{77.7}&\underline{77.9}&\underline{ 77.5}&\underline{74.9}\\
         CoT-Seg with self-correction &Yes&\bf{78.0}&\underline{78.4}&77.4&\underline{76.6}&{\bf 78.2}&{\bf 78.6}&{\bf 77.6}&\underline{74.9}\\
         
         \bottomrule
    \end{tabular}
    }
    \vspace{-0.1in}
    
    \label{tab:cod}
\end{table}

%% file: main.bib
@String(ICLR = {Int. Conf. Learn. Represent.})

@String(ICLR  = {ICLR})

@article{zou2023segment,
  title={Segment everything everywhere all at once},
  author={Zou, Xueyan and Yang, Jianwei and Zhang, Hao and Li, Feng and Li, Linjie and Wang, Jianfeng and Wang, Lijuan and Gao, Jianfeng and Lee, Yong Jae},
  journal={Advances in neural information processing systems},
  volume={36},
  pages={19769--19782},
  year={2023}
}

@article{hurst2024gpt4o,
  title={Gpt-4o system card},
  author={Hurst, Aaron and Lerer, Adam and Goucher, Adam P and Perelman, Adam and Ramesh, Aditya and Clark, Aidan and Ostrow, AJ and Welihinda, Akila and Hayes, Alan and Radford, Alec and others},
  journal={arXiv preprint arXiv:2410.21276},
  year={2024}
}

@article{lai2023lisa,
  title={LISA: Reasoning Segmentation via Large Language Model},
  author={Lai, Xin and Tian, Zhuotao and Chen, Yukang and Li, Yanwei and Yuan, Yuhui and Liu, Shu and Jia, Jiaya},
  journal={arXiv preprint arXiv:2308.00692},
  year={2023}
}

@inproceedings{zou2023generalized,
  title={Generalized decoding for pixel, image, and language},
  author={Zou, Xueyan and Dou, Zi-Yi and Yang, Jianwei and Gan, Zhe and Li, Linjie and Li, Chunyuan and Dai, Xiyang and Behl, Harkirat and Wang, Jianfeng and Yuan, Lu and others},
  booktitle={Proceedings of the IEEE/CVF conference on computer vision and pattern recognition},
  pages={15116--15127},
  year={2023}
}

@inproceedings{liu2023gres,
  title={Gres: Generalized referring expression segmentation},
  author={Liu, Chang and Ding, Henghui and Jiang, Xudong},
  booktitle={Proceedings of the IEEE/CVF conference on computer vision and pattern recognition},
  pages={23592--23601},
  year={2023}
}

@inproceedings{liang2023open,
  title={Open-vocabulary semantic segmentation with mask-adapted clip},
  author={Liang, Feng and Wu, Bichen and Dai, Xiaoliang and Li, Kunpeng and Zhao, Yinan and Zhang, Hang and Zhang, Peizhao and Vajda, Peter and Marculescu, Diana},
  booktitle={Proceedings of the IEEE/CVF conference on computer vision and pattern recognition},
  pages={7061--7070},
  year={2023}
}

@article{liu2025seg,
  title={Seg-zero: Reasoning-chain guided segmentation via cognitive reinforcement},
  author={Liu, Yuqi and Peng, Bohao and Zhong, Zhisheng and Yue, Zihao and Lu, Fanbin and Yu, Bei and Jia, Jiaya},
  journal={arXiv preprint arXiv:2503.06520},
  year={2025}
}

@article{liu2025visionreasoner,
      title={VisionReasoner: Unified Visual Perception and Reasoning via Reinforcement Learning}, 
      author={Yuqi Liu and Tianyuan Qu and Zhisheng Zhong and Bohao Peng and Shu Liu and Bei Yu and Jiaya Jia},
      year={2025},
      journal={arXiv preprint arXiv:2505.12081}
}

@article{snell2024scaling,
  title={Scaling llm test-time compute optimally can be more effective than scaling model parameters},
  author={Snell, Charlie and Lee, Jaehoon and Xu, Kelvin and Kumar, Aviral},
  journal={ICLR},
  year={2025}
}

@article{chen2017deeplab,
  title={Deeplab: Semantic image segmentation with deep convolutional nets, atrous convolution, and fully connected crfs},
  author={Chen, Liang-Chieh and Papandreou, George and Kokkinos, Iasonas and Murphy, Kevin and Yuille, Alan L},
  journal={IEEE transactions on pattern analysis and machine intelligence},
  volume={40},
  number={4},
  pages={834--848},
  year={2017},
  publisher={IEEE}
}

@article{krahenbuhl2011efficient,
  title={Efficient inference in fully connected crfs with gaussian edge potentials},
  author={Kr{\"a}henb{\"u}hl, Philipp and Koltun, Vladlen},
  journal={Advances in neural information processing systems},
  volume={24},
  year={2011}
}

@inproceedings{dias2019semantic,
  title={Semantic segmentation refinement by monte carlo region growing of high confidence detections},
  author={Dias, Philipe Ambrozio and Medeiros, Henry},
  booktitle={Computer Vision--ACCV 2018: 14th Asian Conference on Computer Vision, Perth, Australia, December 2--6, 2018, Revised Selected Papers, Part II 14},
  pages={131--146},
  year={2019},
  organization={Springer}
}

@article{yu2015multi,
  title={Multi-scale context aggregation by dilated convolutions},
  author={Yu, Fisher and Koltun, Vladlen},
  journal={arXiv preprint arXiv:1511.07122},
  year={2015}
}

@article{liu2015parsenet,
  title={Parsenet: Looking wider to see better},
  author={Liu, Wei and Rabinovich, Andrew and Berg, Alexander C},
  journal={arXiv preprint arXiv:1506.04579},
  year={2015}
}

@inproceedings{cheng2022masked,
  title={Masked-attention mask transformer for universal image segmentation},
  author={Cheng, Bowen and Misra, Ishan and Schwing, Alexander G and Kirillov, Alexander and Girdhar, Rohit},
  booktitle={Proceedings of the IEEE/CVF conference on computer vision and pattern recognition},
  pages={1290--1299},
  year={2022}
}

@inproceedings{he2017mask,
  title={Mask r-cnn},
  author={He, Kaiming and Gkioxari, Georgia and Doll{\'a}r, Piotr and Girshick, Ross},
  booktitle={Proceedings of the IEEE international conference on computer vision},
  pages={2961--2969},
  year={2017}
}

@article{badrinarayanan2017segnet,
  title={Segnet: A deep convolutional encoder-decoder architecture for image segmentation},
  author={Badrinarayanan, Vijay and Kendall, Alex and Cipolla, Roberto},
  journal={IEEE transactions on pattern analysis and machine intelligence},
  volume={39},
  number={12},
  pages={2481--2495},
  year={2017},
  publisher={IEEE}
}

@inproceedings{zhao2017pyramid,
  title={Pyramid scene parsing network},
  author={Zhao, Hengshuang and Shi, Jianping and Qi, Xiaojuan and Wang, Xiaogang and Jia, Jiaya},
  booktitle={Proceedings of the IEEE conference on computer vision and pattern recognition},
  pages={2881--2890},
  year={2017}
}

@inproceedings{cheng2020panoptic,
  title={Panoptic-deeplab: A simple, strong, and fast baseline for bottom-up panoptic segmentation},
  author={Cheng, Bowen and Collins, Maxwell D and Zhu, Yukun and Liu, Ting and Huang, Thomas S and Adam, Hartwig and Chen, Liang-Chieh},
  booktitle={Proceedings of the IEEE/CVF conference on computer vision and pattern recognition},
  pages={12475--12485},
  year={2020}
}

@inproceedings{kirillov2019panoptic,
  title={Panoptic segmentation},
  author={Kirillov, Alexander and He, Kaiming and Girshick, Ross and Rother, Carsten and Doll{\'a}r, Piotr},
  booktitle={Proceedings of the IEEE/CVF conference on computer vision and pattern recognition},
  pages={9404--9413},
  year={2019}
}

@inproceedings{kirillov2023segment,
  title={Segment anything},
  author={Kirillov, Alexander and Mintun, Eric and Ravi, Nikhila and Mao, Hanzi and Rolland, Chloe and Gustafson, Laura and Xiao, Tete and Whitehead, Spencer and Berg, Alexander C and Lo, Wan-Yen and others},
  booktitle={Proceedings of the IEEE/CVF international conference on computer vision},
  pages={4015--4026},
  year={2023}
}

@inproceedings{ke2023segment,
 author = {Ke, Lei and Ye, Mingqiao and Danelljan, Martin and liu, Yifan and Tai, Yu-Wing and Tang, Chi-Keung and Yu, Fisher},
 booktitle = {Advances in Neural Information Processing Systems},
 editor = {A. Oh and T. Naumann and A. Globerson and K. Saenko and M. Hardt and S. Levine},
 pages = {29914--29934},
 publisher = {Curran Associates, Inc.},
 title = {Segment Anything in High Quality},
 volume = {36},
 year = {2023}
}

@inproceedings{lai2024lisa,
  title={Lisa: Reasoning segmentation via large language model},
  author={Lai, Xin and Tian, Zhuotao and Chen, Yukang and Li, Yanwei and Yuan, Yuhui and Liu, Shu and Jia, Jiaya},
  booktitle={Proceedings of the IEEE/CVF Conference on Computer Vision and Pattern Recognition},
  pages={9579--9589},
  year={2024}
}

@inproceedings{xia2024gsva,
  title={Gsva: Generalized segmentation via multimodal large language models},
  author={Xia, Zhuofan and Han, Dongchen and Han, Yizeng and Pan, Xuran and Song, Shiji and Huang, Gao},
  booktitle={Proceedings of the IEEE/CVF Conference on Computer Vision and Pattern Recognition},
  pages={3858--3869},
  year={2024}
}

@article{zhang2023next,
  title={Next-chat: An lmm for chat, detection and segmentation},
  author={Zhang, Ao and Yao, Yuan and Ji, Wei and Liu, Zhiyuan and Chua, Tat-Seng},
  journal={arXiv preprint arXiv:2311.04498},
  year={2023}
}

@inproceedings{he2024multi,
  title={Multi-modal instruction tuned llms with fine-grained visual perception},
  author={He, Junwen and Wang, Yifan and Wang, Lijun and Lu, Huchuan and He, Jun-Yan and Lan, Jin-Peng and Luo, Bin and Xie, Xuansong},
  booktitle={Proceedings of the ieee/cvf conference on computer vision and pattern recognition},
  pages={13980--13990},
  year={2024}
}

@article{guo2022images,
  title={From images to textual prompts: Zero-shot vqa with frozen large language models},
  author={Guo, Jiaxian and Li, Junnan and Li, Dongxu and Tiong, Anthony Meng Huat and Li, Boyang and Tao, Dacheng and Hoi, Steven CH},
  journal={arXiv preprint arXiv:2212.10846},
  year={2022}
}

@article{lian2023llm,
  title={Llm-grounded diffusion: Enhancing prompt understanding of text-to-image diffusion models with large language models},
  author={Lian, Long and Li, Boyi and Yala, Adam and Darrell, Trevor},
  journal={arXiv preprint arXiv:2305.13655},
  year={2023}
}

@article{hu2023look,
  title={Look before you leap: Unveiling the power of gpt-4v in robotic vision-language planning},
  author={Hu, Yingdong and Lin, Fanqi and Zhang, Tong and Yi, Li and Gao, Yang},
  journal={arXiv preprint arXiv:2311.17842},
  year={2023}
}

@inproceedings{wei2022chain,
  title={Chain-of-thought prompting elicits reasoning in large language models},
  author={Wei, Jason and Wang, Xuezhi and Schuurmans, Dale and Bosma, Maarten and Xia, Fei and Chi, Ed and Le, Quoc V and Zhou, Denny and others},
  booktitle={Advances in neural information processing systems},
  volume={35},
  pages={24824--24837},
  year={2022}
}

@article{wang2022self,
  title={Self-consistency improves chain of thought reasoning in language models},
  author={Wang, Xuezhi and Wei, Jason and Schuurmans, Dale and Le, Quoc and Chi, Ed and Narang, Sharan and Chowdhery, Aakanksha and Zhou, Denny},
  journal={arXiv preprint arXiv:2203.11171},
  year={2022}
}

@inproceedings{zhang2022automaticchainthoughtprompting,
      title={Automatic Chain of Thought Prompting in Large Language Models}, 
      author={Zhuosheng Zhang and Aston Zhang and Mu Li and Alex Smola},
      year={2023},
      booktitle={International Conference on Learning Representation},
}

@inproceedings{lyu2023faithful,
  title={Faithful chain-of-thought reasoning},
  author={Lyu, Qing and Havaldar, Shreya and Stein, Adam and Zhang, Li and Rao, Delip and Wong, Eric and Apidianaki, Marianna and Callison-Burch, Chris},
  booktitle={The 13th International Joint Conference on Natural Language Processing and the 3rd Conference of the Asia-Pacific Chapter of the Association for Computational Linguistics (IJCNLP-AACL 2023)},
  year={2023}
}

@article{kojima2023largelanguagemodelszeroshot,
      title={Large Language Models are Zero-Shot Reasoners}, 
      author={Takeshi Kojima and Shixiang Shane Gu and Machel Reid and Yutaka Matsuo and Yusuke Iwasawa},
      year={2022},
      journal={arXiv preprint arXiv:2205.11916},
}

@article{mondal2024kamcotknowledgeaugmentedmultimodal,
      title={KAM-CoT: Knowledge Augmented Multimodal Chain-of-Thoughts Reasoning}, 
      author={Debjyoti Mondal and Suraj Modi and Subhadarshi Panda and Rituraj Singh and Godawari Sudhakar Rao},
      year={2024},
     journal={arXiv preprint arXiv:2401.12863},
}

@article{zhang2024multimodalchainofthoughtreasoninglanguage,
      title={Multimodal Chain-of-Thought Reasoning in Language Models}, 
      author={Zhuosheng Zhang and Aston Zhang and Mu Li and Hai Zhao and George Karypis and Alex Smola},
      year={2023},
      journal={arXiv preprint arXiv:2302.00923},
}

@article{lu2022learn,
  title={Learn to explain: Multimodal reasoning via thought chains for science question answering},
  author={Lu, Pan and Mishra, Swaroop and Xia, Tanglin and Qiu, Liang and Chang, Kai-Wei and Zhu, Song-Chun and Tafjord, Oyvind and Clark, Peter and Kalyan, Ashwin},
  journal={Advances in Neural Information Processing Systems},
  volume={35},
  pages={2507--2521},
  year={2022}
}

@inproceedings{mitra2024compositional,
  title={Compositional chain-of-thought prompting for large multimodal models},
  author={Mitra, Chancharik and Huang, Brandon and Darrell, Trevor and Herzig, Roei},
  booktitle={Proceedings of the IEEE/CVF Conference on Computer Vision and Pattern Recognition},
  pages={14420--14431},
  year={2024}
}

@inproceedings{suris2023vipergpt,
  title={Vipergpt: Visual inference via python execution for reasoning},
  author={Sur{\'\i}s, D{\'\i}dac and Menon, Sachit and Vondrick, Carl},
  booktitle={Proceedings of the IEEE/CVF International Conference on Computer Vision},
  pages={11888--11898},
  year={2023}
}

@inproceedings{fan2020camouflaged,
  title={Camouflaged object detection},
  author={Fan, Deng-Ping and Ji, Ge-Peng and Sun, Guolei and Cheng, Ming-Ming and Shen, Jianbing and Shao, Ling},
  booktitle={Proceedings of the IEEE/CVF conference on computer vision and pattern recognition},
  pages={2777--2787},
  year={2020}
}

@inproceedings{kazemzadeh2014referitgame,
  title={Referitgame: Referring to objects in photographs of natural scenes},
  author={Kazemzadeh, Sahar and Ordonez, Vicente and Matten, Mark and Berg, Tamara},
  booktitle={Proceedings of the 2014 conference on empirical methods in natural language processing (EMNLP)},
  pages={787--798},
  year={2014}
}

@inproceedings{luo2020multi,
  title={Multi-task collaborative network for joint referring expression comprehension and segmentation},
  author={Luo, Gen and Zhou, Yiyi and Sun, Xiaoshuai and Cao, Liujuan and Wu, Chenglin and Deng, Cheng and Ji, Rongrong},
  booktitle={Proceedings of the IEEE/CVF Conference on computer vision and pattern recognition},
  pages={10034--10043},
  year={2020}
}

@inproceedings{ding2021vision,
  title={Vision-language transformer and query generation for referring segmentation},
  author={Ding, Henghui and Liu, Chang and Wang, Suchen and Jiang, Xudong},
  booktitle={Proceedings of the IEEE/CVF international conference on computer vision},
  pages={16321--16330},
  year={2021}
}

@inproceedings{wang2022cris,
  title={Cris: Clip-driven referring image segmentation},
  author={Wang, Zhaoqing and Lu, Yu and Li, Qiang and Tao, Xunqiang and Guo, Yandong and Gong, Mingming and Liu, Tongliang},
  booktitle={Proceedings of the IEEE/CVF conference on computer vision and pattern recognition},
  pages={11686--11695},
  year={2022}
}

@inproceedings{yang2022lavt,
  title={Lavt: Language-aware vision transformer for referring image segmentation},
  author={Yang, Zhao and Wang, Jiaqi and Tang, Yansong and Chen, Kai and Zhao, Hengshuang and Torr, Philip HS},
  booktitle={Proceedings of the IEEE/CVF conference on computer vision and pattern recognition},
  pages={18155--18165},
  year={2022}
}

@article{yao2025mmreasonopenendedmultimodalmultistep,
      title={MMReason: An Open-Ended Multi-Modal Multi-Step Reasoning Benchmark for MLLMs Toward AGI}, 
      author={Huanjin Yao and Jiaxing Huang and Yawen Qiu and Michael K. Chen and Wenzheng Liu and Wei Zhang and Wenjie Zeng and Xikun Zhang and Jingyi Zhang and Yuxin Song and Wenhao Wu and Dacheng Tao},
      year={2025},
      journal={arXiv preprint arXiv:2506.23563}, 
}

@article{zhao2025boostingllmreasoningspontaneous,
      title={Boosting LLM Reasoning via Spontaneous Self-Correction}, 
      author={Xutong Zhao and Tengyu Xu and Xuewei Wang and Zhengxing Chen and Di Jin and Liang Tan and Yen-Ting and Zishun Yu and Zhuokai Zhao and Yun He and Sinong Wang and Han Fang and Sarath Chandar and Chen Zhu},
      year={2025},
      journal={arXiv preprint arXiv:2506.06923}, 
}

@article{he2025selfcorrectionrefinementlearningframework,
      title={Self-Correction is More than Refinement: A Learning Framework for Visual and Language Reasoning Tasks}, 
      author={Jiayi He and Hehai Lin and Qingyun Wang and Yi Fung and Heng Ji},
      year={2025},
      journal={arXiv preprint arXiv:2410.04055}, 
}

@article{komeili2021internetaugmenteddialoguegeneration,
      title={Internet-Augmented Dialogue Generation}, 
      author={Mojtaba Komeili and Kurt Shuster and Jason Weston},
      year={2021},
      journal={arXiv preprint arXiv:2107.07566},
}

@article{lewis2021retrievalaugmentedgenerationknowledgeintensivenlp,
      title={Retrieval-Augmented Generation for Knowledge-Intensive NLP Tasks}, 
      author={Patrick Lewis and Ethan Perez and Aleksandra Piktus and Fabio Petroni and Vladimir Karpukhin and Naman Goyal and Heinrich Küttler and Mike Lewis and Wen-tau Yih and Tim Rocktäschel and Sebastian Riedel and Douwe Kiela},
      year={2021},
journal={arXiv preprint arXiv:2005.11401},
}
